\documentclass[letterpaper]{article} 
\usepackage{aaai25}  
\usepackage{times}  
\usepackage{helvet}  
\usepackage{courier}  
\usepackage[hyphens]{url}  
\usepackage{graphicx} 
\usepackage{natbib}  
\usepackage{caption} 
\usepackage{algorithm}
\usepackage{algorithmic}

\usepackage{url}            
\usepackage{booktabs}       
\usepackage{amsfonts}       
\usepackage{nicefrac}       
\usepackage{microtype}      
\usepackage{xcolor}         

\usepackage{sidecap}
\usepackage{dsfont}
\usepackage{amsmath}
\usepackage{xspace}
\usepackage{amsthm}
\usepackage{subfig}
\usepackage[inkscapelatex=false]{svg}
\usepackage{tabularx, booktabs}

\theoremstyle{definition}
\newtheorem{definition}{Definition}[section]

\newtheorem{theorem}{Theorem}

\usepackage{makecell}
\usepackage{pifont}
\newcommand{\cmark}{\textcolor{teal}{\ding{51}}}%
\newcommand{\xmark}{\textcolor{red}{\ding{55}}}%

\newcommand{\ourmethod}{\textsc{NCC}\xspace} 
\newcommand{\ourmethodFullName}{\underline{N}eural \underline{C}onformal \underline{C}ontrol for Time Series Forecasting\xspace}

\newcommand{\flu}{\texttt{flu}\xspace}
\newcommand{\covid}{\texttt{covid-19}\xspace}
\newcommand{\weather}{\texttt{weather}\xspace}
\newcommand{\smd}{\texttt{smd}\xspace}
\newcommand{\electric}{\texttt{electric}\xspace}

\newcommand{\alphabar}{\alpha}

\usepackage{newfloat}
\usepackage{listings}
\DeclareCaptionStyle{ruled}{labelfont=normalfont,labelsep=colon,strut=off} 
\floatstyle{ruled}
\newfloat{listing}{tb}{lst}{}
\floatname{listing}{Listing}
\title{Neural Conformal Control for Time Series Forecasting}

\author {
    Ruipu Li,
    Alexander Rodr{\'\i}guez
}
\affiliations {
    University of Michigan\\
    liruipu@umich.edu, alrodri@umich.edu
}

\usepackage{bibentry}

\begin{document}

\maketitle

\begin{abstract}
We introduce a neural network conformal prediction method for time series that enhances adaptivity in non-stationary environments. 
Our approach acts as a neural controller designed to achieve desired target coverage, leveraging auxiliary multi-view data with neural network encoders in an end-to-end manner to further enhance adaptivity. Additionally, our model is designed to enhance the consistency of prediction intervals in different quantiles by integrating monotonicity constraints and leverages data from related tasks to boost few-shot learning performance. Using real-world datasets from epidemics, electric demand, weather, and others, we empirically demonstrate significant improvements in coverage and probabilistic accuracy, and find that our method is the only one that combines good calibration with consistency in prediction intervals.
\end{abstract}

\begin{links}
    \link{Project page}{https://github.com/complex-ai-lab/ncc}
\end{links}

\section{Introduction}
Quantifying uncertainty in predictions is crucial for various practical applications. Rigorous frameworks are necessary to assess the reliability of these predictions and inform high-stakes decisions, such as hospital resource allocation during epidemics or planning energy supply for cities~\cite{gawlikowski2023survey}. Among these frameworks, conformal prediction (CP)~\cite{vovk2005algorithmic,shafer2008tutorial} stands out for its ability to provide a theoretical coverage guarantee under mild distribution assumptions. However, the guarantee often breaks down in time series forecasting due to temporal dependencies and distribution shifts inherent in time series data. To address these challenges, multiple methods have been proposed, each making specific assumptions about the time series, such as stationarity~\cite{oliveira2022split}, dependence structures~\cite{SPCI}, or exchangeable data and forecasting strategies~\cite{CF-RNNs}. Another line of work~\cite{gibbs2021ACI}, which we refer to as `conformal control', circumvents those assumptions by adopting a control perspective in an online setting, where prediction intervals dynamically adapt to coverage errors to achieve a desired target coverage. Based on this, several variations have been introduced to refine the controller logic and enhance adaptivity~\cite{ConformalPID,zaffran2022adaptivecp,bhatnagar2023improvedaci}.

Despite these advancements, significant gaps remain between current CP methods and the demands of real-world applications. For instance, the epidemic forecasting initiatives hosted by the US Centers for Disease Control and Prevention (CDC) in response to diseases such as Ebola~\cite{viboud2018rapidd,johansson2019open}, influenza~\cite{reich2019collaborative,mathis2024evaluation}, and COVID-19~\cite{cramer2022evaluation} highlight these challenges. Because these predictions inform critical policymaking decisions, the CDC requires forecasters to provide a comprehensive view of future possibilities. Consequently, submissions must include different confidence levels that adherence to \emph{distributional consistency}~\cite{reich2019collaborative,cramer2022evaluation}, which prohibits crossing prediction intervals. Failure to meet these standards makes forecasts invalid for CDC acceptance.
Conformal control methods do not guarantee this because of disjoint calibration for each confidence level and our experiments revealed that some of them fail to provide valid forecasts as low as 0\% of cases.
In addition, forecasting teams are encouraged to utilize any relevant datasets, and previous work has demonstrated the effectiveness of leveraging \emph{multi-view datasets} from heterogeneous sources~\cite{rodriguez2021deepcovid,qian2020and} and across multiple modalities~\cite{deng2020cola,kamarthi2021camul}. 
For example, distribution shifts driven by human behavior can be anticipated by analyzing digital surveys and mobility data~\cite{rodriguez2024machine}.
However, CP methods often underutilize the auxiliary information, which could better inform adaptivity mechanisms and improve calibration.
Moreover, high-stakes applications like epidemic forecasting require readiness to address emerging outbreaks, such as recent cases of H5N1 and measles, where data is scarce. This underscores the need to enhance the \emph{few-shot learning} capabilities of current CP methods.

In response to these challenges, we propose integrating recent ideas from conformal control with advances in deep learning. Our method, \ourmethodFullName (\ourmethod), incorporates control-inspired loss functions to enable the neural network to function as a predictive controller. This approach offers a flexible, end-to-end learning framework for conformal prediction using neural networks without losing the theoretical guarantee on coverage. We summarize our main contributions as follows. 
\begin{enumerate} 
    \item \textbf{Neural Control for Conformal Prediction:} 
    We introduce \ourmethod, a conformal control method designed to harness the capabilities of neural networks to capture patterns that inform adaptivity to changes in time series. To achieve this, we develop multiple neural modules to identify these patterns and incorporate loss functions that enhance control and improve the efficiency of prediction intervals.
    \item \textbf{Enabling Consistent Prediction Intervals and Few-Shot Learning:} 
    We leverage the flexibility of \ourmethod to constrain our neural network to yield consistent prediction intervals through a process involving a monotonicity loss and a test-time adaptation (TTA) procedure. Additionally, our neural architecture is designed to facilitate transfer learning across multiple tasks (e.g., regions) and enhance few-shot learning (FSL) performance.
    \item \textbf{Extensive Experiments and Theoretical Guarantees:} We evaluate \ourmethod on a variety of real-world datasets and against other state-of-the-art CP methods. We show that \ourmethod achieves superior empirical performance while maintaining theoretical long-term coverage guarantees.
\end{enumerate} 
More details in our contributions are outlined in Table~\ref{tab:compare}. 

\section{Preliminaries}
\label{sec:background}

Here we provide background on CP methods in time series, highlighting their limitations in real-world applications, which motivate our proposed method.

\par\noindent\textbf{Conformal Prediction (CP) Setup.}
Let us introduce the basic setup for conformal predictions in time series forecasting. We focus our attention on an online learning setting where feature-target pairs are observed sequentially~\cite{gibbs2021ACI}. At each time step $t$, we have a univariate response $y_t \in \mathcal{Y}$, features $X_t$ ($X_t$ may include $y_t$), and use a forecasting model 
to generate predictions $\hat{y}_{t+\tau}$ for future time steps, where $\tau\in\mathbb{N}^{+}$ is the forecasting horizon.
A user shall specify one or several target error rate $\alphabar\in(0, 1)$ (also known as miscoverage rate). Then, we will refer to a method as well-calibrated in uncertainty if it constructs prediction interval $\hat{C}_{t}$ such that $y_{t}$, the actual future value of the response, falls within $\hat{C}_{t}$ at least $100(1-\alphabar)\%$ of the time. 

A CP method typically involves splitting the observed data into a training set, used to fit the forecasting model, and a calibration set used for creating the prediction sets~\cite{papadopoulos2007conformal}. A non-conformity score \( s_t(y_t, \hat{y}_t) \) is defined to measure how closely the observed value \( y_t \) `conforms' with the predictions from the fitted model. For example, one of the most widely-used and simple non-conformity score in regression problems is \( s_t(y_t, \hat{y}_t) = |y_t - \hat{y}_t| \), which we adopt in this paper. Then, within the calibration set, we calculate the non-conformity scores and determine \( q_t^{\alpha} \), which is the \( (1-\alpha) \) quantile of these scores. In conventional CP settings, where all data points are assumed to be exchangeable, the prediction interval $\hat{C}_{t}^\alpha = \{y\in \mathcal{Y}: s_t(y,\hat{y_{t}}) \leq q_{t}^{\alpha}\}$ is guaranteed to include \( y_t \) with a probability of at least \( 1-\alpha \)~\cite{vovk2005algorithmic}. 

\par\noindent\textbf{CP in Time Series.}
In time series forecasting, the exchangeability assumption does not hold. As a result, a coverage guarantee at time \( t \) can only be achieved with trivial methods that produce prediction intervals of infinite size~\cite{ConformalPID}. The most feasible outcome is achieving a valid long-term coverage, defined as \( \frac{1}{T}\sum_{t=1}^T (\mathrm{err}^{\alpha}_t - \alpha) < o(1) \). Here $\mathrm{err}^{\alpha}_t$ is the coverage error at time step $t$ indicating that the ground truth data is not covered by the prediction interval associated with $\alpha$. This error is defined as \( \mathrm{err}^{\alpha}_t := \mathds{1}(s_t > \hat{q}_t^{1-\alphabar}) \), where \( \mathds{1} \) is the indicator function. \( o(1) \) denotes a quantity that approaches zero as \( T \rightarrow \infty \).
Such guarantees can be achieved using an adaptive error rate \( \alpha_t \)~\cite{gibbs2021ACI}, different from the constant target error rate \( \alpha \) specified by the user. 
\( \alpha_t \) evolves over time according to the update rule: \( \alpha_{t+1} = \alpha_t + \eta (\alpha - \mathrm{err}^\alpha_t) \), where \( \eta \) is a learning rate. This approach allows \( \alpha_t \) to adjust dynamically based on the observed error \( \mathrm{err}^\alpha_t \), enhancing the adaptivity of the prediction intervals. Recent work has recast the task of adaptivity as a control problem with a feedback loop~\cite{ConformalPID,zaffran2022adaptivecp,bhatnagar2023improvedaci} and achieved notable performance improvements maintaining theoretical guarantees on long-term coverage. As mentioned earlier, we refer to these approaches as `conformal control' methods. 
\begin{table}[!t]
\centering
\caption{Comparison with other time series CP methods. Conformal control methods include ACI~\cite{gibbs2021ACI} and variants~\cite{ConformalPID, zaffran2022adaptivecp, DtACI}. Methods with dist. assumptions include~\cite{CF-RNNs, nexcp, oliveira2022split}.}
\resizebox{0.98\linewidth}{!}{
\begin{tabularx}{\linewidth}{l >{\centering\arraybackslash}X >{\centering\arraybackslash}X >{\centering\arraybackslash}X}
     & \makecell{\smash{\hspace{-2.0cm}\raisebox{0cm}{\rotatebox{-20}{Conformal control}}}} 
     & \makecell{\smash{\hspace{-2.0cm}\raisebox{0cm}{\rotatebox{-20}{Dist. assumptions}}}}
     & \makecell{\smash{\hspace{-2.0cm}\raisebox{0cm}{\rotatebox{-20}{\ourmethod (this paper)}}}} \vspace{0.80cm}\\ 
\midrule
Control-based adaptivity & \cmark & \xmark & \cmark \\ 
Coverage guarantees & \cmark & \cmark & \cmark \\ 
Incorporates multi-view data & \xmark & \xmark & \cmark \\ 
Consistent prediction intervals & \xmark & \cmark & \cmark \\ 
Data efficiency for FSL & \xmark & \xmark & \cmark \\ 
End-to-end learning & \xmark & \xmark & \cmark \\
\end{tabularx}
}
\label{tab:compare}
\end{table}
\par\noindent\textbf{Practical Limitations in Conformal Control.}
One key limitation of existing conformal control methods is the lack of mechanisms to ensure adaptivity of prediction intervals. As exemplified by the update rule above, the adaptivity of these methods relies solely on adjustments based on observed coverage errors, potentially overlooking other important patterns, such as recurring trends in non-conformity scores. Additionally, these methods have not yet explored how to incorporate multi-view auxiliary data into their adaptive updates.
Another issue is the inconsistency among prediction intervals for a given set of error rates. In standard CP methods, the error rate $\alpha$ is fixed, making the non-conformity score--the $(1 - \alpha)$ quantile of calibration set scores--naturally monotonic with respect to $\alpha$~\cite{shafer2008tutorial}. However, conformal control methods such as ACI~\cite{gibbs2021ACI} and variants~\cite{ConformalPID, DtACI, zaffran2022adaptivecp}, and some ML-based CP methods~\cite{SPCI}, where the quantile of non-conformity scores are constantly adapted, the non-conformity score is not necessarily monotonic. This can lead to the overlapping and crossing of prediction intervals, a problem known as quantile crossing. We formalize this property with the following definition, which comes naturally from the non-decreasing property of the quantile function.

\par\noindent\textbf{Definition 1.} (Distributional Consistency)
Given the $n$ prediction intervals $(\hat{y}^{\alpha_1}_{lower}, \hat{y}^{\alpha_1}_{upper}), \ldots, (\hat{y}^{\alpha_n}_{lower}, \hat{y}^{\alpha_n}_{upper})$ and a set of $n$ error rates $\mathbf{A} := \alpha_1, \ldots, \alpha_n$, we say the prediction intervals are consistent if, for any $i, j \in [1, n]$, where $\alpha_i < \alpha_j$, it holds that $\hat{y}^{\alpha_i}_{lower} \leq \hat{y}^{\alpha_j}_{lower}$ and $\hat{y}^{\alpha_i}_{upper} \geq \hat{y}^{\alpha_j}_{upper}$.

\par\noindent\textbf{Connection to Related Works.}
As mentioned earlier, due to temporal distribution shifts, using the quantile of scores in the calibration set no longer gives “conformal” non-conformity scores on test data. Therefore, we can view the efforts made by time series CP methods as proposing better quantile predictors that adapts to the temporal changes. For ACI and variants~\cite{gibbs2021ACI, bhatnagar2023improvedaci, DtACI}, the quantile function evolves across time through changing the error rate. C-PID~\cite{ConformalPID} directly updates the quantile using online gradient descent. SPCI~\cite{SPCI} learns a random forest quantile estimator. Our work follows this line of work by using a neural network. It is worth noting that conformalized quantile regression~\cite{romano2019conformalized} is quite different. CQR and other quantile regression methods aim to learn a conditional quantile function on the target variable Y~\cite{SPCI}. Thus, they fit the quantile regressor on Y directly, while ours fits the quantile regressor on user-defined non-conformity scores. More on related work can be found in our Appendix.

\section{Our Approach}
\ourmethod consists of a predictive controller for coverage errors based on neural networks, control-inspired losses, and a test-time adaptation procedure. \ourmethod enables:
\textbf{(1)} \emph{Improved adaptivity through representation learning and multi-view data:} Leveraging a neural network allows us to exploit patterns beyond error rates and seamlessly incorporate multi-view data, which provides valuable information for adapting prediction intervals. Our architecture is designed with encoders that incorporate appropriate inductive biases and control-inspired losses to effectively learn to function as a predictive controller.
\textbf{(2)} \emph{Improved consistency of prediction intervals without sacrificing performance}: We design our neural network architecture and loss functions to ensure that predicted quantiles are monotonically decreasing (i.e., non-increasing) with respect to $\alpha$. Although our prediction intervals are constructed sequentially, an adjustment step required for coverage guarantees (known as the `conformalization step') may disrupt this property. To address this issue, we introduce a monotonicity loss during training and a test-time adaptation (TTA) procedure to preserve this property during inference.
\textbf{(3)} \emph{Guaranteed long-term coverage through a differentiable conformalization step}: While the prediction intervals produced by our neural network are well-calibrated and generally consistent, they do not inherently provide the theoretical coverage guarantees offered by other CP methods. By incorporating an adjustment term to these prediction intervals, we can conformalize them, thereby preserving the theoretical coverage guarantees of traditional conformal control methods.
\begin{figure}[t]
    \centering
    \includegraphics[width=0.98\columnwidth]{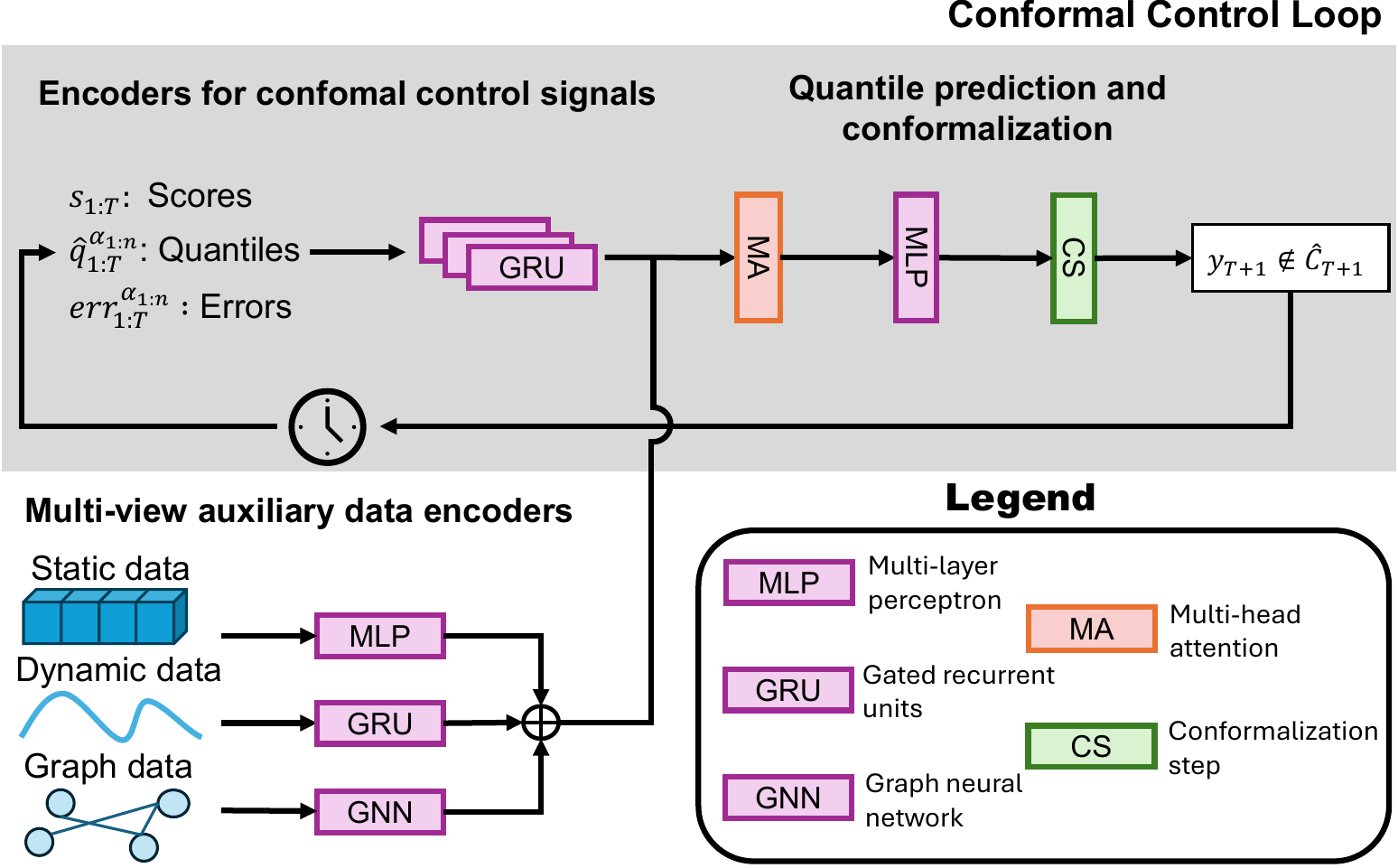}
    \caption{Model architecture of \ourmethod. The shaded area is the conformal control loop of our method. MLP, GRU and GNN are used to encode multi-view data. Additionally, GRUs are employed to encode past errors, predictions, and  scores. These embeddings are integrated using multi-head cross-attention mechanisms and subsequently passed through an MLP to predict the quantiles of scores. Finally, the predicted quantiles are refined through a conformalization step.}
    \label{fig:model arch}
\end{figure}
\subsection{Neural Network Architecture}
Our model consists of two major modules: a conformal control loop and multi-view data encoders. 

\par\noindent\textbf{Conformal Control Loop.} The main part of our model is a conformal control loop as shown in the shaded area in Figure \ref{fig:model arch}. At time $T$, we will have available the conformal control signals $\mathrm{err}_T^{\alpha}$ and $\hat{q}_T^{\alpha}$ for each $\alpha \in \{\alpha_1, \ldots, \alpha_n\}$, and non-conformity score $s_T$. Then, considering their past, we have three conformal signals: $\mathrm{err}_{1:T}^{\alpha} := \{\mathrm{err}_t^{\alpha} | 1 \leq t \leq T\}$, $ \hat{q}_{1:T}^{\alpha} := \{\hat{q}_t^{\alpha} | 1\leq t \leq T\}$ for each $\alpha$ and $s_{1:T} := \{s_t |1 \leq t \leq T\}$. These conformal control signals decide how we adapt our predicted quantiles of the non-conformity scores. Therefore, we encode the three sequences using GRUs~\cite{chung2015gated}. The learned embedding is then dynamically combined using a multi-head attention mechanism. Specifically, we have a combined embedding $z_{\text{combined}} = \text{Multi-head}(z_{\text{data}}, z_{\mathrm{err}}, z_{q}, z_{s})$ which is subsequently fed into an MLP with a ReLU layer as the final layer. We denote the output from the quantile predictor as $\{\Tilde{q}^{\alpha_1}, \ldots, \Delta \Tilde{q}^{\alpha_n}\}$, which holds $\Delta \Tilde{q}^{\alpha_i} \geq 0$ because of the final ReLU layer. Therefore, for two positions $i$ and $j$, we will have $i < j$, $\alpha_i > \alpha_j$ (i.e., $\alpha$'s are sorted in decreasing order).
Note that $\Tilde{q}$ is not yet our final prediction interval $\hat{q}_t^{\alpha}$, but instead an intermediate value. 
$\Tilde{q}$ from the quantile predictor is then conformalized through a conformalization step (we detail this step later in this section). 

\par\noindent\textbf{Multi-view Data Encoders.}
Following previous work in multi-view learning~\cite{yan2021deep}, we use different encoders for each view. A natural way to determine views is based on data modalities, which are typically time series (sequential), graph (relational) and static for forecasting tasks~\cite{kamarthi2021camul}. Within each modality, we find different views (e.g., one for each data source), which can be particularly useful when each view provides a different perspective or has varying noise levels. Let $j$ be the number of data views; then, the combined representation of the multi-view data is $z_{\text{data}} = \text{Multi-head}\big\{g_{\theta_j}(X^{(j)})\big\}$, where $\text{Multi-head}$ is a multi-head attention layer~\cite{vaswani2017attention}, and $g_{\theta_j}$ is a neural network module parameterized by $\theta_j$, chosen based on the modality of view $j$. For example, we use a GRU for views with sequence data, a GCN~\cite{kipf2016semi} to encode relational data, and feed-forward networks for static fixed-sized features. This flexible setup allows us to leverage the representational power of neural networks to transfer knowledge across multiple tasks. Building on previous successes in transfer learning for CP~\cite{fisch2021few}, our framework can facilitate this process, for instance, by incorporating region encoding as one of our static views.

\subsection{Control-inspired Losses for Improved Adaptivity}
\label{section: losses}

To enable a data-driven model to effectively control the coverage of our conformal prediction intervals, we incorporate losses that promote controller-like behavior.

\par\noindent\textbf{Coverage Adaptivity via Quantile Loss.}
To learn a conditional quantile distribution of the nonconformity scores, we use the quantile (pinball) loss. Denote the quantile loss~\cite{romano2019conformalized} at the level $(1-\alpha)$ as $\rho_{1-\alpha}(z)$, which is defined as $(1-\alpha) |z|$ for $z > 0$ and $(\alpha) |z|$ for $z \leq 0$~\cite{koenker2005quantile}. When the predicted interval covers the ground truth data point (i.e., $s_t \leq \hat{q}_t^{\alpha}$), the quantile loss is $(-\alpha)(s_t - \hat{q}_t^{\alpha})$; otherwise, it is $(1-\alpha)(s_t - \hat{q}_t^{\alpha})$. Thus:
\begin{equation}
    \mathcal{L}_Q = \max\big\{(1-\alpha)(s_t - \hat{q}_t^{\alpha}), (-\alpha)(s_t - \hat{q}_t^{\alpha})\big\}, \quad \forall \alpha\in \mathbf{A}.
\end{equation}

\par\noindent\textbf{Coverage Stability via Integrative Coverage Loss.}
Distribution shifts can lead to fluctuating coverage errors (sometimes largely positive, sometimes largely negative). 
These fluctuations can be misleading for decision makers trying to understand the reliability of prediction intervals. Therefore, an important feature for our prediction intervals is not only to adapt but also maintain a stable coverage error, a property we refer to as coverage stability. To address this, we design an integrative \emph{coverage loss} to ensure coverage stability. 

First, let us define the running coverage error ($\Bar{\mathrm{err}}$), which integrates errors over the most recent $w$ time steps. At time $T+1$, this error is given by: $\Bar{\mathrm{err}}^{\alpha}_w(T) := \frac{1}{w}\sum_{t=T-w+1}^{T} \mathrm{err}^{\alpha}_t$. Using this error, we can determine whether our future prediction intervals \underline{need} to cover the ground truth data to achieve the target error rate. To facilitate this, we define the indicator function $\mathrm{cov}^{\alpha}(T)$, which indicates whether coverage is necessary for a prediction interval made at $T$ (remember that at time $T$ we predict for $T+\tau$). Our intuition is the following: if $\Bar{\mathrm{err}}^{\alpha}_w(T) > \alpha$, we want the next prediction interval to be large enough to cover the ground truth, i.e., $\mathrm{err}^{\alpha}_{T+1}$ should be 0. In this case, we set $\mathrm{cov}^{\alpha}(T) = 1$. Otherwise, if $\Bar{\mathrm{err}}^{\alpha}_w(T) \leq \alpha$, the prediction interval should be smaller, i.e., $\mathrm{err}^{\alpha}_{T+1}$ should be 1. In this case, we set $\mathrm{cov}^{\alpha}(T) = 0$. This is:
\[
\mathrm{cov}^{\alpha}(T)  = \begin{cases}
            0, & \Bar{\mathrm{err}}^{\alpha}_w(T) \leq \alpha\\
      1, & \text{otherwise.}
         \end{cases}
\]
We can think of $\mathrm{cov}^{\alpha}(T)$ as a binary label, which should align with the current coverage condition indicated by $1 - \mathrm{err}^{\alpha}_{T+1}$.  Therefore, we choose to define our coverage loss based on the log loss. For simplicity, we denote $\mathrm{err}^{\alpha}_{T+1}$ as $\mathrm{err}^{\alpha}$ and denote $\mathrm{cov}^{\alpha}(T)$ as $\mathrm{cov}^{\alpha}$ in the loss function.
\begin{equation}
\label{equation:coverage_loss}
    \mathcal{L}_C = -(1-\mathrm{cov}^{\alpha})\log(\mathrm{err}^{\alpha}) - \mathrm{cov}^{\alpha}\log(1-\mathrm{err}^{\alpha}), \quad \forall \alpha\in \mathbf{A}.
\end{equation}
To make $\mathrm{err}^{\alpha}_t$ differentiable, we approximate it as: 
\[\mathrm{err}^{\alpha}_t \approx \mathrm{sigmoid}\Big(\frac{s_t - \hat{q}_t^{\alpha}}{K}\Big),
\]
where $\mathrm{sigmoid}(x) = \frac{1}{1+e^{-x}}$ is the sigmoid function, and $K$ is a constant. With an appropriate choice of $K$, $\mathrm{err}^{\alpha}_t$ is close to $1$ when $s_t > \hat{q}_t^{\alpha}$ and close to $0$ otherwise.

\par\noindent\textbf{Designing Efficient Prediction Intervals.} %
Previous work has shown that quantile regression methods trained with quantile loss sometimes produce unnecessarily wide prediction intervals~\cite{romano2019conformalized}. To prevent the prediction intervals from becoming excessively wide, we incorporate a regularization term, the efficiency loss $\mathcal{L}_E$. This term penalizes large prediction interval sizes when the ground truth is covered, thus promoting more efficient intervals.
\begin{equation}
    \mathcal{L}_E = (y_t - \hat{q}_t^{\alpha})^2 \times (1 - p_\mathrm{cov}), \quad \forall \alpha\in \mathbf{A}.
\end{equation}

\subsection{Long-term Coverage Guarantees Through Conformalization} 
The quantiles of nonconformity scores predicted by the neural network do not inherently ensure valid long-run coverage~\cite{romano2019conformalized}. To achieve theoretical guarantees, we further conformalize the outputs of our model as per Algorithm~\ref{alg:algorithm}, line 6. Note that our algorithm only considers the case where we calibrate for one step ahead predictions for simplicity, but it works for any prediction horizon $\tau$. Refer to the Appendix for more details. For some $\alpha$, we will consider the prediction interval constructed as $\hat{C}_{t} = \{y\in \mathcal{Y}: s_t(y_{t},\hat{y_{t}}) \leq (\hat{q}_{t})\}$, where \( s_t(y_t, \hat{y}_t) = |y_t - \hat{y}_t| \). Unlike quantile tracking from C-PID, we do not use the update rule: $q_{t+1} = q_t + \eta(\mathrm{err}_t - \alpha)$. Instead, we replace $\mathrm{err}_t$ with the running average error rate $\frac{1}{w}\sum_{t=T-w+1}^T\mathrm{err}_t$. In the setting of C-PID, since $\mathrm{err}_t$ is either 1 or 0, $q_{t+1} \neq q_t$. This means that the prediction intervals predicted by the model are always adjusted. Our objective is to make the least adjustments to the prediction intervals. Therefore, instead of using error at $t$, we use the average of errors in a window of size $w$. Then when the prediction intervals are well calibrated, the average is close to $\alpha$ and the prediction interval would not be changed significantly.

\begin{algorithm}[tb]
\caption{Forward pass over \ourmethod at time step $T$.}
\label{alg:algorithm}
\textbf{Parameter}: User-specified window size $w$, learning rate $\eta$, and a set of $n$ decreasing target error rates $\mathbf{A} := \alpha_1, \cdots, \alpha_n$. \\
\textbf{Input}: Historical data, along with past non-conformative scores, coverage errors and quantiles, which we group and denote as $\mathcal{D} := \{X_t, y_t, s_t, \{\mathrm{err}^{\alpha}_t, \hat{q}^{\alpha}_t\}_{\alpha \in \mathbf{A}} \}_{t=1}^T$. Quantile predictor $\mathcal{M}$.
\\
\textbf{Output}: Conformalized prediction intervals $\hat{C}_{T+1}^{\alpha}, \forall \alpha \in \mathbf{A}$.

\begin{algorithmic}[1]
\STATE For all $\alpha \in \mathbf{A}$, initialize $\Delta \Tilde{q}^{\alpha}_1 = 0$.
\STATE Forward pass over quantile predictor: $\Tilde{q}^{\alpha_1}_{T+1}, \ldots, \Tilde{q}^{\alpha_n}_{T+1} = \mathcal{M}(\mathcal{D})$
\FORALL{$i \in 1,\ldots,n$}
    \STATE Calculate the average error rate for window $\Bar{\mathrm{err}}^{\alpha_i}_w (T) = \frac{1}{w}\sum_{t=T-w+1}^T\mathrm{err}^{\alpha_i}_t$
    \STATE Update $\Delta \Tilde{q}^{\alpha_i}$ as $\Delta \Tilde{q}^{\alpha_i}_{T+1} = \Delta \Tilde{q}^{\alpha_i}_{T} + \eta(\Bar{\mathrm{err}}^{\alpha}_w (T) - \alpha_i)$
    \STATE Conformalization step: $\hat{q}^{\alpha_i}_{T+1} = \Tilde{q}^{\alpha_i}_{T+1} + \Delta \Tilde{q}^{\alpha_i}_{T+1}$
    \IF{forward pass is used at test time (inference)}
        \STATE TTA: Train learnable vector $h\in \mathbb{R}^{n}$ (until desired DCS) so that $\hat{q}^{\alpha_i}_{T+1} + h_i$ monotonically increases as $\alpha_i$ decreases.
    \ENDIF
    \STATE $\hat{C}_{T+1}^{\alpha_i} = \{y | s_t(\hat{y}_{T+1}, y) \leq \hat{q}^{\alpha_i}_{T+1}\}$
\ENDFOR
\end{algorithmic}
\end{algorithm}
\begin{theorem}
\label{thm:long_term}
(Long-term coverage~\cite{ConformalPID})
    For sufficiently large $T$, assume that $\Delta q \leq b$ for some $b \in \mathds{R}$. Then the prediction interval $\hat{C}_t$ constructed by our method \ourmethod satisfies
$ \frac{1}{T}\sum_{t=1}^T \mathds{1}(y_t \notin \hat{C}_t) = \alpha + o(1) $.
\end{theorem}
See proof of this theorem in our Appendix\ref{sec:proof}.

\par\noindent\textbf{Remark}: The predicted quantiles from the neural network lack a theoretical guarantee of long-term coverage. However, as demonstrated in the proof of \textbf{Theorem 1}, applying the aforementioned algorithm conformalizes the predicted quantiles. This ensures that \ourmethod possesses the same long-term coverage guarantee as ACI and C-PID.

\subsection{Improved Distributional Validity Through Monotonic Quantiles}
\label{section: valid}
Quantile crossing leads to invalid prediction intervals. To address this issue, we first construct the prediction intervals cumulatively. We use a ReLU layer as the final layer of the quantile predictor. As noted before, we denote the output of the quantile predictor as $\Tilde{q}^{\alpha_1}, \ldots, \Delta \Tilde{q}^{\alpha_n}$, where $\Delta \Tilde{q}^{\alpha_i} \geq 0$ and for $i < j$, $\alpha_i > \alpha_j$. 
Note that $\Tilde{q}^{\alpha_1}$ is inherently non-negative, as our chosen absolute residue score function constrains the nonconformity scores to be non-negative. If a different score function is employed, other requirements on the nonconformity scores can be similarly implemented.
Then we cumulatively sum the output: $\Tilde{q}^{\alpha_k} = \Tilde{q}^{\alpha_1} + \sum_{i=2}^{k} \Delta \Tilde{q}^{\alpha_i}, k \in [2, n]$.
Although the cumulative design ensures that $\Tilde{q}$ is monotonic with respect to $\alpha$, the conformalization step disrupts this consistency by adding an adjustment term. Thus, similar to~\cite{rodriguez2023einns}, we apply a monotonicity loss to $\hat{q}$:
$$L_M(\hat{q}_t^{\alpha}, \alpha) = \max\Big(0, \frac{\partial \hat{q}_t^{\alpha}}{\partial \alpha}\Big)$$ 
We approximate the derivatives using a finite difference scheme, which makes our monotonicity loss take the form:
$$ L_M = \sum_{i=1}^{n-1} \max\Big(0, \frac{\hat{q}_t^{\alpha_{i+1}} - \hat{q}_t^{\alpha_i}}{\alpha_{i+1} - \alpha_i} \Big)$$
Such loss encourages the model to produce monotonic output, i.e, when $\alpha_1 > \alpha_2$, $\hat{q}^{\alpha_1} < \hat{q}^{\alpha_2}$. 

At test time, while our cumulative design and monotonicity loss enhance overall monotonic behavior, there are instances where distribution validity is compromised due to few non-monotonic terms. To address this, we apply a TTA procedure right before conformalization. Specifically, we pass the combined embedding from the model through a MLP layer, which outputs a term $h_i$ for each $\alpha_i, i \in [1, n]$. For each $i$, we adjust $\Tilde{q}^{\alpha_i}$ by adding $h_i$ and then apply the monotonicity loss to the conformalized $\hat{q}$s. If the DCS (i.e., the number of instances where the model provides consistent quantiles) meets the user-specified threshold, we return the prediction and proceed to the next time step. Otherwise, we repeat the adaptation process until the desired DCS is achieved.

\section{Experiments}
\begin{figure*}[!htb]
  \includegraphics[width=\linewidth]{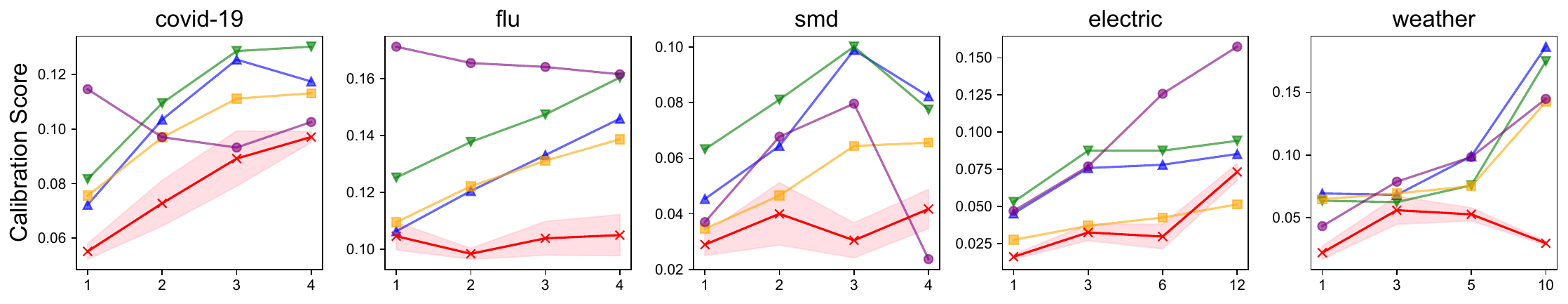}
  \includegraphics[width=\linewidth]{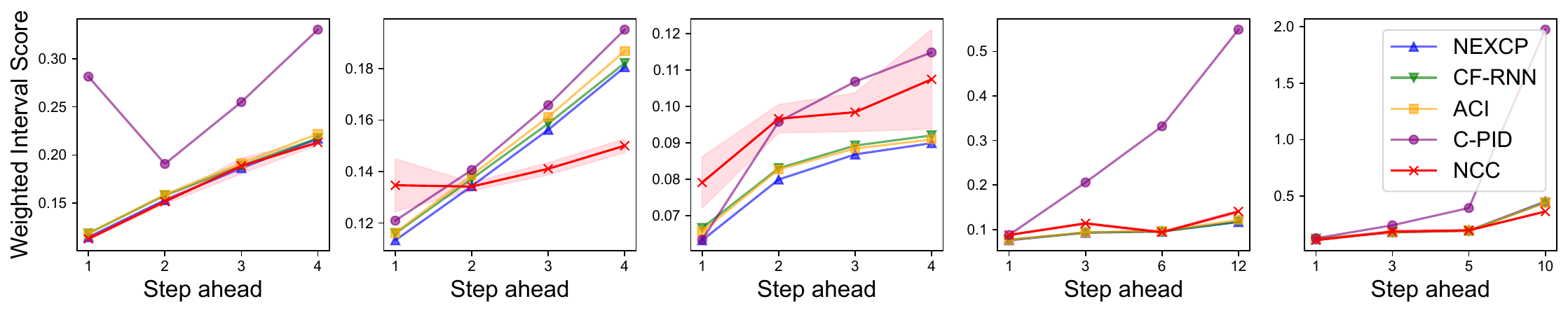}
  \caption{Comparison of CS and WIS across five datasets for four different steps ahead predictions. Our method (\ourmethod) is highlighted with a red line, with the pink-shaded area around it indicating the error margin. \ourmethod has the best CS across all datasets. Except for the \smd dataset, \ourmethod consistently achieves the best or near-best performance in WIS.
  }\label{figure: wis_results}
\end{figure*}
We comprehensively compare \ourmethod with popular CP time series methods across multiple datasets and base forecasters. To align with real-world applications, all experiments emulate a real-time forecasting setup. Additionally, we conduct experiments on few-shot learning and model ablation. Detailed information about our experimental setup and results is provided in the Appendix.
\subsection{Experimental Setting}
\par\noindent\textbf{Datasets and Base Forecasters.}  We evaluate on a diverse set of publicly available real-world datasets, including commonly-used datasets in time series forecasting (\weather, \electric) and datasets that demonstrate significant amount of temporal distribution shifts (\flu, \covid, \smd). 
To illustrate that our method does not depend on a specific base forecaster, we get the point predictions using three base forecasters: seq2seq GRU model, Theta model~\cite{assimakopoulos2000theta}, and Informer model~\cite{zhou2021informer}.

\par\noindent\textbf{Baselines.}  We compare the calibration power of \ourmethod to four popular time series conformal prediction methods.
Conformal forecasting RNNs (CF-RNN)~\cite{CF-RNNs} generate prediction intervals for multi-horizon forecasts while assuming data exchangeability, similar to conventional conformal prediction methods. Nonexchangeable conformal prediction (NEXCP)~\cite{nexcp} introduces weighted quantiles to deal with distribution shifts. ACI~\cite{gibbs2021ACI}, adaptive conformal inference, which adapts the user-specified error rate using online gradient descent.
C-PID~\cite{ConformalPID}, conformal PID control, which introduces PID control to time series CP. Since using a scorecaster (D-part) in C-PID breaks the theoretical coverage guarantee, we only use the quantile tracker and error integrator, which are the P and I part in PID.
As is shown in Table \ref{tab:compare}, CF-RNN and NEXCP are methods under distribution shift assumptions. ACI and C-PID are conformal control methods.
Note that these baselines cannot incorporate auxiliary data. Therefore, to provide a fair comparison, \ourmethod uses multi-view data only when studying the effect of incorporating multi-view data.

\par\noindent\par\noindent\textbf{Evaluation Metrics.}  
We evaluate our model and baselines using carefully chosen metrics that are widely used in the literature to measure probabilistic accuracy and calibration~\cite{kamarthi2021camul,pmlr-v139-xu21h_enbpi}. These metrics include Calibration Score (CS), Continuous Ranked Probability Score (CRPS) and Weighted Interval Score (WIS). The CS is used to measure how much the empirical coverage rate differs from the ideal coverage rate. CRPS evaluates the efficiency and coverage at multiple miscoverage levels. WIS is often used as an approximation of CRPS~\cite{bracher2021evaluating}. In our experiments, the results for CRPS and WIS are similar. Thus, we present the CRPS in the appendix. For CS, WIS and CRPS, smaller values indicate better preformance.
To evaluate distribution consistency, we propose the Distribution Consistency Score (DCS), which quantifies the ratio of consistent predictions to the total number of predictions. 

\subsection{Comparison with Other CP Methods}
\label{section:cmp}

\begin{figure*}[!htb]
\minipage{0.27\textwidth}
  \centering
  \includegraphics[width=\linewidth]{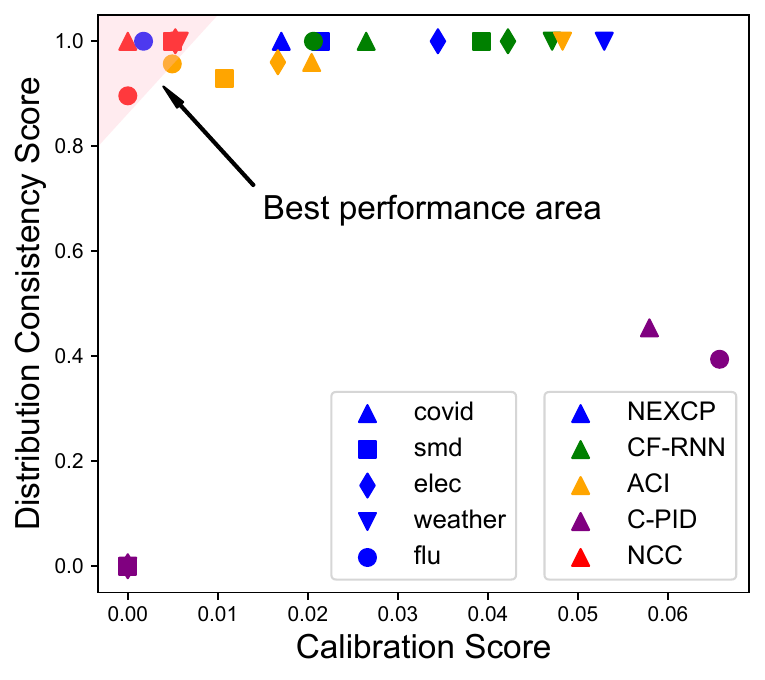}
  \text{(a)}
\endminipage\hfill
\minipage{0.27\textwidth}
  \centering
  \includegraphics[width=\linewidth]{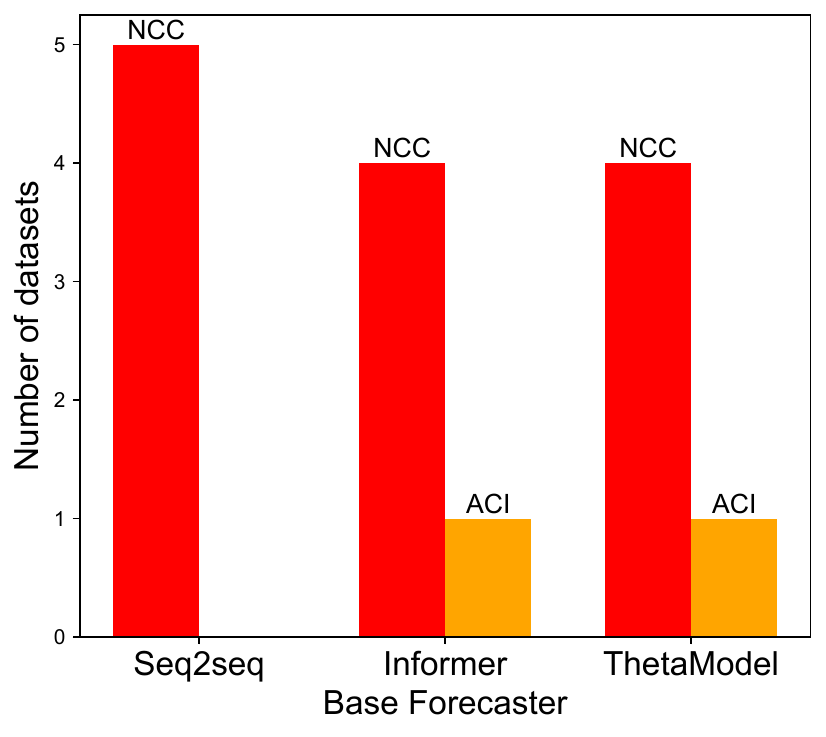}
  \text{(b)}
\endminipage\hfill
\minipage{0.27\textwidth}
  \centering
  \includegraphics[width=\linewidth]{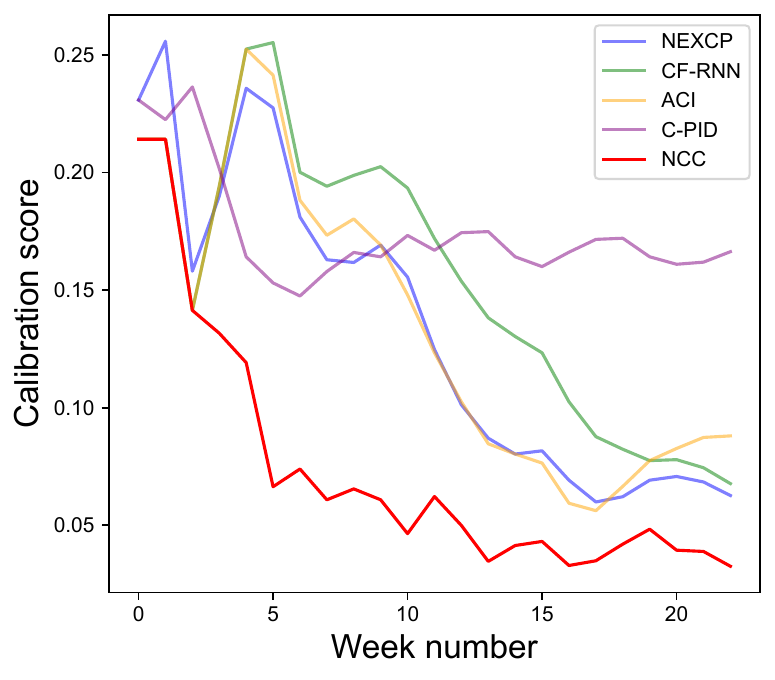}
  \text{(c)}
\endminipage
\minipage{0.1666\textwidth}
  \centering
  \includegraphics[width=\linewidth]{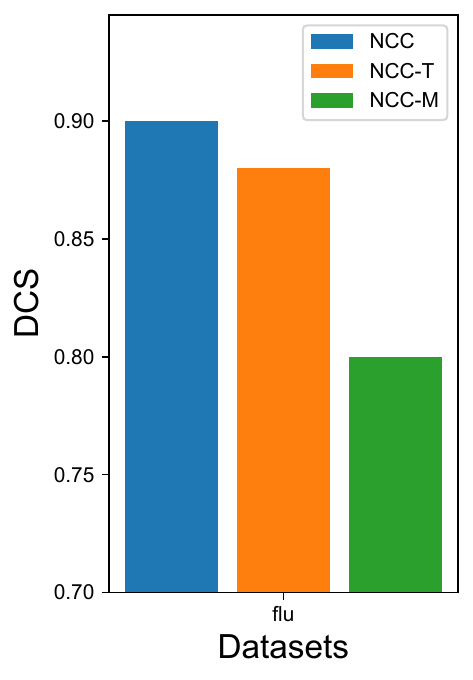}
  \text{(d)}
\endminipage
\caption{(a) DCS versus CS, using the \textbf{unsorted} results. The top left corner represents the optimal performance region. (b) The number of datasets where our method achieves the best CS. The methods that do not achieve the best CS are not shown in this plot. (c) Cumulative average CS on \covid in a few-shot learning setup. (d) Ablation study on TTA and monotonicity loss. Comparison of \ourmethod, \ourmethod without TTA (\ourmethod-T) and \ourmethod without TTA and monotonicity loss (\ourmethod-M).}
\label{figure: combined}
\end{figure*}
Here we focus on the seq2seq model as our base forecaster. Additional results with other base forecasters are provided in our Appendix. As discussed before, most time series CP methods exhibit some degree of distribution inconsistency, which makes intervals invalid. For example, C-PID has 0\% valid intervals for \flu. To enable comparisons, we first make the intervals valid by sorting them, which is a commonly used approach~\cite{chernozhukov2010quantile, chernozhukov2009improving}.

Figure \ref{figure: wis_results} provides a summary of our main findings, with detailed numerical values available in Table \ref{table: numerical results} in the Appendix. The plot illustrates the CS and WIS across five datasets, evaluated at four different forecasting horizons. For CS, \ourmethod consistently achieves the smallest values, indicating superior calibration performance, except for the four-step-ahead cases on \smd and \electric. Notably, CS tends to increase as the step becomes larger, yet \ourmethod demonstrates stability, maintaining a nearly flat trend even as prediction accuracy decreases at longer horizons. For WIS, \ourmethod outperforms the baselines on the \covid, \flu, and \weather datasets. On \electric, \ourmethod performs comparably to the best results, showcasing its overall effectiveness.
It might be surprising that ACI outperforms C-PID in many cases in CS. This occurs because C-PID, though it adapts quickly under distribution shifts, produces less consistent intervals. Then the sorting greatly reduces the performance of the C-PID. In contrast, \ourmethod, which also exhibits strong adaptivity, maintains good performance even when sorting is applied. Figure \ref{figure: combined}(a) illustrates the reason. The figure shows DCS against CS when the prediction intervals are unsorted. It reveals that, except for our method, the more adaptive a method is (indicated by a lower CS), the more inconsistent its predictions tend to be (indicated by a lower DCS). For example, while C-PID achieves the best CS on most datasets, it shows nearly zero DCS for \smd, \electric, and \weather. \ourmethod, however, successfully finds a balance between adaptivity and distribution consistency. Across all five datasets, \ourmethod's markers are consistently near the top left corner of the plot, which represents the optimal performance region. 

\subsection{Few-shot Learning Ability}
\label{section:fewshot}
In this section, we demonstrate that our method achieves superior performance in a few-shot learning setup. For testing, we used COVID-19 hospitalization data from the 2021-2022 respiratory virus season within the \covid dataset, training on the data from the 10 weeks before the respiratory virus season. The test period spans 28 weeks. Figure \ref{figure: combined}(c) shows the cumulative average calibration score starting from the sixth week. During the respiratory virus season, COVID-19 hospitalization exhibits a rapid increase followed by a decrease, making calibration particularly challenging at the start and end of the season.
As shown by the red line in the plot, despite the aforementioned challenges, the CS of \ourmethod rapidly drops below 0.1, demonstrating its strong adaptability. In contrast, the CS of other methods remains high by week 10. Furthermore, \ourmethod consistently achieves the lowest calibration scores. Notably, the baselines either require a number of observations of non-conformity scores before they can adapt, or rely on online gradient descent for adjustments, resulting in significantly high initial CS. Since \ourmethod does not rely on online gradient descent and effectively uses additional data from other regions through transfer learning, it achieves optimal calibration performance from the beginning.

\subsection{Robust Calibration across Various Base Forecasters}
In this section, we demonstrate that \ourmethod consistently achieves the best performance, irrespective of the base forecaster used. In addition to the seq2seq model, we further use an Informer and a Theta model as the base forecaster. Detailed results are provided in the Appendix in Table \ref{table: other base forecaster}. Figure \ref{figure: combined}(b) presents the number of datasets in which each method excels, based on CS. Across all base forecasters, our method outperforms others by achieving the best CS on at least four datasets, which highlights its superior calibration capabilities.

\subsection{Ablation Study}
\label{sec:ablation study}
\par\noindent\textbf{Effect of Quantile Loss and Coverage Loss.}
We evaluate across all five datasets on one step ahead predictions when either quantile loss or coverage loss is excluded. Note that when coverage loss is removed, we also exclude efficiency loss. The results, presented in Table \ref{table:ablation study remove loss} in the Appendix, show a significant decrease in CS when coverage loss is excluded, which suggests that quantile loss alone is insufficient to achieve a valid empirical coverage rate. However, quantile loss remains essential to the model as it enables learning the conditional distribution, which serves as a foundation for coverage loss to further improve performance. When quantile loss is removed, the CS degrades only slightly, but the WIS increases dramatically, indicating that the resulting prediction intervals become much larger (less efficient).

\par\noindent\textbf{Effect of Incorporating Monotonicity Loss and TTA.}
In this section, we explore the methods used to enhance the distribution consistency of prediction intervals. In Figure \ref{figure: combined}(d) we compare the Distribution Consistency Score (DCS) for \ourmethod, \ourmethod without TTA (\ourmethod-T), and \ourmethod without both TTA and monotonicity loss (\ourmethod-M). The results demonstrate that both the monotonicity loss and TTA contribute to improving distribution consistency.

\par\noindent\textbf{Effect of Incorporating Multi-view Data.}
We compare the performance of \ourmethod with and without the use of multi-view data. Table \ref{table:auxilary info} in Appendix summarizes the results. Overall, incorporating multi-view data reduces CS by up to 18\%.

\section{Conclusion}
We introduce \ourmethod, one of the first end-to-end deep learning frameworks for time series CP. By leveraging the powerful data ingestion capabilities and data-driven flexibility of deep learning models, our method provides more adaptive prediction intervals showcasing better calibration. An advantage of our approach is its ability to incorporate domain expert requirements, such as avoiding quantile crossing, which is a common challenge in real-world applications. Through extensive evaluations on diverse datasets and comparisons against competitive baselines, we demonstrate the superior performance and practicality of \ourmethod. We believe that our work paves the way for a tighter integration of conformal prediction techniques with deep learning, enabling more reliable and trustworthy decision-making processes in various domains that rely on accurate time series forecasting and UQ.

\par\noindent\textbf{Limitations and Future Work:} We acknowledge some limitations of our method. In a few datasets, our method constructs wider prediction intervals than others. The reason can be that our model is trained using multiple losses for different objectives, in which case the efficiency loss can be overlooked due to its scale. We also found our model’s performance can be sensitive to hyperparameters.
Future work could explore advanced methods for balancing losses and making architectural modifications to enhance robustness. Additionally, integrating recent advancements in time series prediction, such as decomposition techniques~\cite{zeng2023transformers} and generative models~\cite{yuan2024diffusionts}, may offer further improvements.

\section*{Acknowledgements}
This work was supported in part by the Centers for Disease Control and Prevention Award NU38FT000002 and start-up funds from the University of Michigan.

\bibliographystyle{aaai24}

\newpage
\appendix

\section{Proof of theoretical coverage}
\label{sec:proof}
\textbf{Theorem 1}: For sufficiently large T, assume that $\Delta q \leq b$ for some $b \in \mathds{R}$. Then the prediction interval $\hat{C}_t$ constructed following the algorithm above satisfies
$$ \frac{1}{T}\sum_{t=1}^T \mathds{1}(y_t \notin \hat{C}_t) = \alpha + o(1) $$
\textbf{Proof.} Applying the update rule iteratively, we have,
\begin{eqnarray}
\Delta q_{T+1}&=& \Delta q_T + \eta(\Bar{\mathrm{err}}_T - \alpha) \nonumber    \\
~&=& \Delta q_{T-1} + \eta(\Bar{\mathrm{err}}_T - \alpha) + \eta(\Bar{\mathrm{err}}_{T-1} - \alpha) \nonumber \\
~&=& \cdots \nonumber \\
~&=& \Delta q_1 + \sum_{i=1}^T \eta (\Bar{\mathrm{err}}_i - \alpha)
\end{eqnarray}
For sufficiently large T, 
\begin{eqnarray}
\label{equation: approx}
\sum_{i=1}^T \Bar{\mathrm{err}}_i&=& \sum_{i=1}^T (\frac{1}{w}\sum_{t=i-w}^i\mathrm{err}_t) \nonumber    \\
~&\approx& w\times \frac{1}{w} \sum_{i=1}^T \mathrm{err}_i \nonumber \\
~&=& \sum_{i=1}^T \mathrm{err}_i \nonumber \\
\end{eqnarray}
Here is an intuitive explanation of the approximation in Equation \ref{equation: approx}. When calculating the average error rate in a window of size $w$, most $err_t$ (More exactly, when $t \geq w$ and $t \leq T-w$) is calculated $w$ times and divided by $w$, which gives $err_t$. Since T is large enough, we can ignore the errors at the beginning and end of the this sequence.
Then, 
\begin{eqnarray}
\label{equation: p1}
\Delta q_{T+1}&=& \Delta q_1 + \sum_{i=1}^T \eta (\Bar{\mathrm{err}}_i - \alpha) \nonumber    \\
~&=& \Delta q_1 + \sum_{i=1}^T \eta (\mathrm{err}_i - \alpha) \nonumber \\
\end{eqnarray}
Rearrange Equation \ref{equation: p1} gives:
$$ \frac{1}{T} \sum_{i=1}^T\mathrm{err}_i = \alpha + \frac{\Delta q_{T+1} - \Delta q_1}{\eta T} $$
Also note that $\Delta q_1 = 0$ and $\Delta q_{T+1} \leq b$ for some $b \in \mathds{R}$, then 
$$ \frac{1}{T} \sum_{i=1}^T\mathrm{err}_i = \alpha + o(1) $$
which proves the above theorem. Here $o(1)$ is a quantity that approaches zero as $T \rightarrow \infty$.

\section{Related Work}

While deep neural networks (DNNs) exhibit strong performance in predicting time series, their complexity, driven by a large number of parameters, challenges the understanding of their outputs. Occasionally, DNNs may produce completely inaccurate predictions without user awareness, which can be especially harmful in high-stakes applications. Recognizing this, there is a growing focus on incorporating UQ methods to enhance the interpretability and reliability of predictions. 

\paragraph{Deep Uncertainty Quantification}
A number of deep UQ methods have been developed while few focus on time series settings although the latter is more common in practical applications. Bayesian RNN methods quantify uncertainty by modeling the distribution of parameters\cite{fortunato2017bayesian, mcdermott2019bayesian}. However, it usually involves modification to the fundamental forecasting model. Many non-parametric deep learning methods are also proposed for UQ~\cite{epifnp, HQPI, SQR-RNN}. Deep ensemble methods are another kind of methods used for uncertainty quantification \cite{lakshminarayanan2017simpleensemblemodel, ensemblemodel1}. Deep ensembles use multiple models that are trained on different datasets or initialized with different weights to generate many predictions. The uncertainty is estimated based on the variance of these predictions. From Bayesian RNNs to deep ensembles, these methods either requires substantial modification to the base forecasting model or needs a lot of additional training and predicting cost to quantify the uncertainty. 

\paragraph{Conformal Prediction}
Conformal Prediction (CP) has emerged as a model-agnostic approach in recent years, supported by well-defined mathematical proofs that establish theoretical coverage guarantees \cite{shafer2008tutorial}. By defining an error rate $\alpha$, CP generates either a set of potential predictions (for classification) or a prediction interval (for regression), encompassing the true prediction with a probability of $1 - \alpha$. CP has been widely used to address uncertainty quantification (UQ) in deep neural networks (DNNs). In contrast to the previously mentioned methods, CP stands out for its independence from the underlying forecasting model's details. It requires no prior knowledge of the base forecasting model and does not necessitate any modifications to it; it solely utilizes the model's predictions. This distinctive property makes CP versatile and applicable to a diverse range of regression and classification problems. 

In time series, the assumption of exchangeability is challenged by potential temporal distribution shifts.
To address this issue, several Conformal Prediction (CP) methods tailored to adapt to temporal distribution shifts have been proposed. One such method is Conformal Forecasting Recurrent Neural Networks (CF-RNNs), which is designed for generating prediction intervals in the context of multi-horizon time series forecasts \cite{CF-RNNs}. CF-RNNs assumes that the time series data is still exchangeable. Based on this, it predicts the prediction intervals for each horizon separately using standard conformal prediction method. While among the first to apply the conformal prediction framework to time series settings, CF-RNNs is no different from standard CP methods. Adaptive conformal inference (ACI) is the first method that modifies the standard conformal prediction framework to account for changes in online settings \cite{gibbs2021ACI}. Rather than relying on a fixed error rate ($\alpha$), ACI employs a dynamic approach by updating $\alpha$ over time using a pinball loss. In practical terms, if the preceding prediction interval successfully encompasses the target $y_{T+1}$, the error rate is increased, leading to a more tightly defined subsequent prediction interval. Conversely, if the last prediction interval fails to capture the target, the error rate is decreased, resulting in a wider prediction interval in the next time step. One notable and intriguing property of ACI is its capability to adjust to temporal distribution shifts while concurrently ensuring a guarantee on long-term coverage. Specifically, for a sufficiently lengthy time series, it can be proved that the empirical coverage rate converges to $1 - \alpha$. Following ACI, several additional conformal prediction methods have been introduced \cite{zaffran2022adaptivecp,bhatnagar2023improvedaci}. Building upon the foundation laid by ACI, these methods improve the update logic for increased effectiveness. Another noteworthy conformal prediction method that addresses uncertainty modeling and adapts to temporal distribution shifts is Conformal PID Control (Conformal PID) \cite{ConformalPID}.

Conformalized Quantile Regression (CQR) utilizes quantile regression for generating conformity scores \cite{romano2019conformalized}. Sequential Predictive Conformal Inference (SPCI) leverages a quantile regression model to exploit potential temporal dependencies among non-conformity scores, forecasting the next non-conformity score \cite{SPCI}. Distributional Conformal Prediction (DCP) models the conditional distribution $Y_t | X_t$ to enable adaptive prediction intervals based on the most recent input data \cite{chernozhukov2021distributional}. To improve the adaptivity to distribution shifts, we provide as much information to our model as possible, including the input features to the base forecasting model, previous non-conformity scores and previous error rates, so that the model can capture the uncertainty from the input features and model the possible temporal relationship in the non-conformity scores and error rates.

\section{More Details on Distributional Consistency}

The transition from standard CP methods to time series CP methods introduces issues in distributional consistency, an issue also known as quantile crossing. In standard CP methods, the error rate $\alpha$ is fixed, making the non-conformity score (the $1 - \alpha$ quantile of calibration set scores) naturally monotonic with respect to $\alpha$~\cite{shafer2008tutorial}. However, in ACI, conformal PID, and some ML-based CP methods, where the error rate changes or deep learning models are introduced, the non-conformity score is not necessarily monotonic, leading to the overlapping and crossing of prediction intervals, also known as quantile crossing. For example in ACI, consider $\alpha_1$ and $\alpha_2$, $\alpha_1 < \alpha_2$. At time t, assume that ${\alpha_1}_t < {\alpha_2}_t$, and the prediction interval of $\alpha_1$ covers while $\alpha_2$ does not. Then $\mathrm{err}_t^{\alpha_1} = 0$ and $\mathrm{err}_t^{\alpha_2} = 1$. Then ${\alpha_1}_{t+1} = {\alpha_1}_t + \eta(\alpha_1)$, ${\alpha_2}_{t+1} = {\alpha_2}_t + \eta(\alpha_2 - 1)$. Then
\begin{equation}
    \label{equation:qc_aci}
    \eta(\alpha_1 - \alpha_2 + 1) > {\alpha_2}_t - {\alpha_1}_t \implies {\alpha_1}_{t+1} > {\alpha_2}_{t+1}
\end{equation}
Therefore at $t+1$, if $\eta(\alpha_1 - \alpha_2 + 1) > {\alpha_2}_t - {\alpha_1}_t$, ${\alpha_1}_{t+1}$ will be greater than ${\alpha_2}_{t+1}$, which means that the prediction interval of $\alpha_1$ will be smaller than $\alpha_2$. In real-world applications, when $\alpha_1$ and $\alpha_2$ are close and $\eta$ is relatively large, the condition is easy to meet. In conformal PID, because of error integrator, the adjustments to quantiles of non-conformity scores are even more common and aggressive, which leads to even more frequent quantile crossing. 

\begin{definition}[Distributional Consistency--DC]
    \label{def:dc}
    Given the prediction intervals $(\hat{y}^{\alpha_1}_{lower}, \hat{y}^{\alpha_1}_{upper}), \cdots, (\hat{y}^{\alpha_n}_{lower}, \hat{y}^{\alpha_n}_{upper})$ associated with respective error rates $\alpha_1, \cdots, \alpha_n$, we say the prediction intervals are consistent if for any $i, j \in [1, n]$, if $\alpha_i < \alpha_j$, $\hat{y}^{\alpha_i}_{lower} \leq \hat{y}^{\alpha_j}_{lower}$ and $\hat{y}^{\alpha_i}_{upper} \geq \hat{y}^{\alpha_j}_{upper}$.
\end{definition}
This definition comes naturally from the non-decreasing property of the quantile function.
Consider a prediction interval constructed using the non-conformity function \( s_t = |y_t - \hat{y}_t| \). Then for some $\alpha$, the prediction interval is $[\hat{y} - \hat{q}^{\alpha}, \hat{y} + \hat{q}^{\alpha}]$. Then since the quantile function is non-decreasing, for two error rates $\alpha_1 < \alpha_2$, $\hat{q}^{\alpha_1} \geq \hat{q}^{\alpha_2}$. Then we have that the prediction interval associated with $\alpha_1$ is at least as wide as that of $\alpha_2$. 
Then according to the definition of the quantile function,
\begin{equation}
    \label{equation: def quantile}
    Q(x) := inf\{ y\in \mathds{R}: \int \mathds{1} (f(u)\leq y ) du \geq x \}
\end{equation}
The cumulative distribution function (CDF) follows $F(x) = \int_{-\infty}^xPr[Y=y]dy$. Since for any $y$, $Pr[Y=y] \geq 0$, $F(x)$ is monotonically increasing. Then we can define the inverse CDF as $Quantile(p) := F^{-1}(p)$.

\subsection{More on TTA}
\begin{figure}[ht]
    \centering
    \includegraphics[width=0.98\columnwidth]{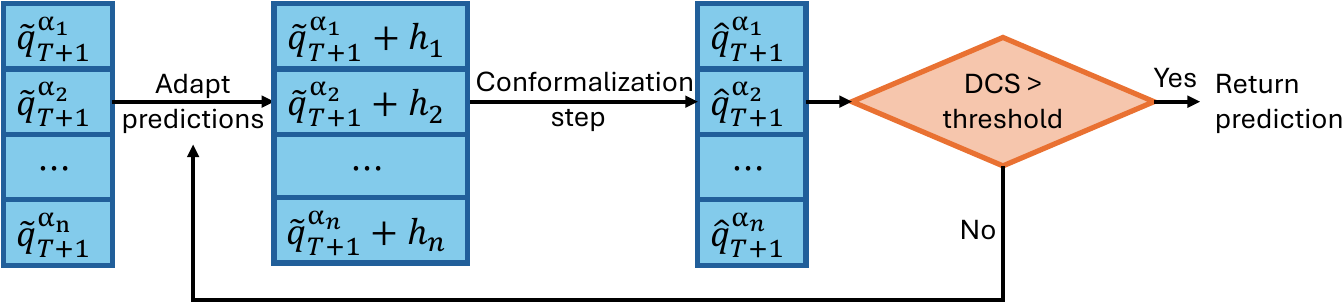}
    \caption{TTA procedure used at inference time to further enhance prediction interval consistency.}
    \label{fig:tta} 
\end{figure}

Figure \ref{fig:tta} shows the TTA process.

\section{Experimental Details}
\label{appendix:expdetails}

\subsection{Experimental Setup}
We conducted our experiments using NVIDIA A40 GPUs on a Linux system. Each experiment took approximately three hours to complete. Training for 100 epochs takes around 10 minutes. The time for inference is negligible. In each experiment, we first trained the model on the training data. During the testing phase, the model was retrained at every $\Delta t$ time step, where $\Delta t$ is a hyperparameter. The duration of one epoch's training varied across datasets, with approximately twenty epochs requiring two to three minutes of training time.

The evaluation of the forecasting results follows a similar approach to that used in CDC forecasting tasks. We constructed prediction intervals for eleven different confidence levels. We then evaluated the Calibration Score, Cumulative Ranked Probability Score, Weighted Interval Score, and coverage rates.

\subsection{More Details on Datasets}
\label{appendix:data}
We use five real-world datasets in the experiments. For each dataset, we train and forecast with our base forecasters.
\noindent\textbf{1. \flu (Flu forecasting):}
We use aggregated and anonymized search counts of flu-related symptoms web searches by Google\footnote{\url{https://pair-code.github.io/covid19_symptom_dataset}} from each US state to predict incidence of influenza 1 week ahead in the future. The flu incidence rate is represented by wILI (weighted Influenza related illness) values released weekly by CDC for 8 HHS regions of USA\footnote{\url{https://predict.cdc.gov/post/5d8257befba2091084d47b4c}}. The \flu dataset has diverse multimodal data useful for predictions. 
1. \textit{Sequence data}: This contains reference points of symptoms features for all previous forecasting seasons for all HHS regions.
2.  \textit{Graph data}: It has an adjacency matrix $A^{(j)}$ encodes the neighbourhood information between HHS regions that share a border. 
3. \textit{Static data}: This includes the current week index and the region.
For \flu, we use the data in respiratory virus season (from week 40 to week 20 in the next year) of 2021-2022, 2022-2023 and 2023-2024.
\noindent\textbf{2. \covid (COVID-19 Hospitalization forecasting):}
Same as \flu, on this dataset, we aggregate the data from Google and weekly released data from CDC. For \covid, we use data in respiratory virus season 2021-2022.
\noindent\textbf{3. \weather:}
\weather includes 21 indicators of weather, such as air temperature, and humidity. Its data is recorded every 10 min for 2020 in Germany \cite{lai2018modeling_datasets}.
\noindent\textbf{4. \electric:}
\electric\cite{SPCI} is a well-establisted dataset for time series forecasting.
\noindent\textbf{5. \smd:}
\smd includes 5 weeks of data from two server machines with 38-dimensional sensor inputs \cite{su2019robust}. The two server machines are selected because of the distribution shifts they demonstrated.

\subsection{More Details on Metrics}
\label{sec:details_metrics}
    $\bullet$ \textbf{Calibration Score (CS)} it is used to measure the overall calibration of a model at multiple error rates~\cite{kuleshov2018accurate,epifnp}. We compute the proportion $k_m(c)$ of prediction distributions that include the actual outcome at each confidence level $c = 1-\alpha$. A perfectly calibrated model has $k_m(c)$ very close to $c$. Therefore, $CS$ is defined as:
\[CS(M) = \int_0^1 |k_m(c)-c| dc.\]
which can be approximated by summation over small intervals.

    \noindent$\bullet$\textbf{Cumulative Ranked Probability Score (CRPS)} is a commonly used metric for assessing probabilistic forecasts, extending mean average error to the realm of probabilistic predictions \cite{gneiting2014probabilistic}. CRPS achieves its minimum value when the predicted distribution matches the true distribution on average, hence evaluating both accuracy and calibration. For a ground truth $y^{(t)}$, let the predicted probability distribution be $\hat{p}_{y^{(t)}}$ with mean $\hat{y}^{(t)}$. Also let $\hat{F}_{y^{(t)}}$ be the CDF. Note that $y^{(t)}$ is not $y_t$ we refer to in other parts. Actually, $y^{(t)}$ is $q_t$. CRPS is defined as:
\[CRPS(\hat{F}_y, y) = \int_{-\infty}^\infty (\hat{F}_y(\hat{y}) - \mathbf{1}\{\hat{y}>y\})^2 d\hat{y}.\]

    \noindent$\bullet$\textbf{Weighted Interval Score (WIS)} is used for forecasts provided in an interval format. It is a weighted sum of the interval score of multiple prediction intervals associated with different confidence levels\cite{bracher2021evaluating}. To define the weighted interval score, we first define the interval score as follows:
    \[
    \begin{aligned}
        \mathrm{IS}_{\alpha}(F, y) &:= (u - l)+ \frac{2(l-y)}{\alpha} \times \mathds{1}(y < l) \\
        & + \frac{2(y - u)}{\alpha} \times \mathds{1}(y > u)
    \end{aligned}
    \]
    where $F$ is the forecast, $l$ is the $\frac{\alpha}{2}$ quantile of $F$, $u$ is the $1 - \frac{\alpha}{2}$ quantile of $F$.\\
    Then for $K$ $\alpha s$, $\alpha_1 > \cdots > \alpha_K$ and the predictive median m, the weighted interval score is defined as:
    \[
    \begin{aligned}
        \mathrm{WIS}_{\alpha_{0:K}}&(F, y) := \frac{1}{K+1/2} \times \\
        & (w_0\times |y - m| + \sum_{k=1}^K (w_k \times \mathrm{IS}_{\alpha_k}(F, y)))
    \end{aligned}
    \]
    where $w_0, \cdots, w_K$ are weights. In this paper, we set $w_0 = \frac{1}{2}$ and $w_k = \frac{\alpha_k}{2},k\in [1, K]$.
    
    A more detailed description can be found in \cite{cramer2022evaluation}.

\subsection{Training Details}
\label{appendix:training}

\textbf{Losses:} As discussed in Section \ref{section: losses}, we use three losses in our training process, quantile loss, coverage loss and efficiency loss.
Our final loss to minimize during training our model is the following:
\begin{equation}
    \label{equation:all_loss_func}
    \mathcal{L} = \lambda_Q \mathcal{L_Q} + \lambda_C \mathcal{L}_C + \lambda_E \mathcal{L}_E,
\end{equation}
where $\lambda_Q$, $\lambda_C$, $\lambda_E$ are weights for our losses, which will be hyperparamters.

\textbf{Training Procedure:} We observed that optimizing quantile loss and coverage loss simultaneously at the beginning often resulted in frequent unsuccessful experiments. This is because quantile loss models the conditional distribution, whereas coverage loss adjusts based on previous coverage rates. To address this, we implemented a three-stage training strategy. In the first stage, we train the model using only quantile loss. In the second stage, we incorporate coverage loss and efficiency loss. Finally, in the third stage, we train with all three losses together. Our approach uses an online learning setup, where the model is retrained as new data becomes available. In the beginning, we pad predictions with zeros and pad errors with ones when they are not available yet. To manage computational costs, we retrain every two steps on \covid and every five steps on the other datasets. During retraining, we continue to apply the three-stage training strategy, but with a reduced number of epochs.

For the \covid and \flu datasets, we train and evaluate our model across 52 regions in the United States as well as on the entire country. To facilitate transfer learning across regions, we incorporate region encodings into our model. Similarly, for multi-horizon forecasting, our method allows for the calibration of all predictions within the horizon using a single model by including the look-ahead step as an additional feature.

\textbf{Hyperparameter Selection:} For the end-to-end conformal model, we optimize the model and training parameters using Bayesian optimization~\cite{frazier2018tutorial}. We perform hyperparameter optimization for 100 iterations on each dataset, which leads to significant performance improvements. This is crucial because certain hyperparameters require careful tuning. For instance, the factor \textbf{K} in the loss function has a substantial impact on performance. If \textbf{K} is too small, the gradient of the sigmoid function becomes nearly zero for most inputs. Conversely, if \textbf{K} is too large, the sigmoid function produces small outputs, resulting in a reduced coverage loss. We apply the same optimization strategy to select hyperparameters for the base forecasters as well.

\section{Additional Experiment Results}
\subsection{Effect of Quantile Loss and Coverage Loss}
Here in Table \ref{table:ablation study remove loss} are the detailed results for the effect of quantile loss and coverage loss.

\begin{table}[ht]
\centering
\caption{Ablation results of \ourmethod. We compare \ourmethod, \ourmethod without quantile loss and \ourmethod without coverage loss.}
\label{table:ablation study remove loss}
\begin{tabular}{@{\extracolsep{8pt}}llcc}
\toprule
\textbf{Dataset}  & \textbf{Method}  & \textbf{CS} &  \textbf{WIS}   \\ 
\midrule
\flu                  & \ourmethod               & 0.104                   & 0.134                         \\
                      & \ourmethod-C          & 0.154                    & 0.128                        \\
                      & \ourmethod-Q          & 0.0967                   & 0.162                        \\ \hline
\covid                & \ourmethod               & 0.0551                   & 0.113                         \\
                      & \ourmethod-C          & 0.0883                   & 0.107                          \\
                      & \ourmethod-Q          & 0.0600                   & 0.218                          \\ \hline
\weather              & \ourmethod               & 0.0221                  & 0.108                      \\ 
                      & \ourmethod-C          & 0.0466                  & 0.110                       \\
                      & \ourmethod-Q          & 0.0235                  & 0.235                          \\\hline
\smd                  & \ourmethod               & 0.0289                   & 0.0791                      \\
                      & \ourmethod-C          & 0.115                    & 0.0779                         \\
                      & \ourmethod-Q          & 0.0444                   & 0.101                           \\\hline
\electric             & \ourmethod               & 0.0161                  &  0.0876                    \\
                      & \ourmethod-C          & 0.0472                  & 0.0820                        \\
                      & \ourmethod-Q          & 0.0175                  & 0.134                          \\
\bottomrule
\end{tabular}
\end{table}

\subsection{Effect of incorporating multi-view data}

As shown in Table \ref{table:auxilary info}, incorporating multi-view data generally leads to improved results. Notably, the multi-view data results in an approximately 10\% improvement in calibration score for \covid. In the cases of \flu and \weather, where the multi-view data does not enhance performance, the results remain similar to those achieved without using this data.

\begin{table}[ht]
\centering
\caption{Effect of incorporating multi-view data. We compare the performance of our method with and without the use of multi-view data. We denote the model with multi-view data as \ourmethod+.}
\label{table:auxilary info}
\resizebox{0.499\textwidth}{!}{
\centering
\begin{tabular}{@{\extracolsep{8pt}}llccc}
\toprule
& & \textbf{Calibration} & \multicolumn{2}{c}{\textbf{Probabilistic Acc.}} \\
\cmidrule{3-3}
\cmidrule{4-5}
\textbf{Dataset}   & \textbf{Method} & \textbf{CS} & \textbf{CRPS}   & \textbf{WIS} \\
\midrule
\flu                  & \ourmethod+               & $0.143 \pm 0.080$        & $0.127 \pm 0.004$ & $0.120\pm 0.003$ \\ 
                      & \ourmethod               & $0.104 \pm 0.004$        & $0.142 \pm 0.010$ & $0.134\pm 0.010$ \\ 
\midrule
\covid                & \ourmethod+           & $0.0501 \pm 0.0049$         & $0.115\pm 0.002$ & $0.112\pm 0.002$ \\
                      & \ourmethod           & $0.0551 \pm 0.0037$         & $0.116\pm 0.002$ & $0.113\pm 0.002$ \\
\midrule
\weather              & \ourmethod+               & $0.0237\pm 0.0086$         & $0.118\pm 0.003$  & $0.114\pm 0.004$ \\
                      & \ourmethod               & $0.0221\pm 0.007$         & $0.115\pm 0.002$  & $0.108\pm 0.001$ \\
\midrule
\smd                  & \ourmethod+           & $0.0244\pm 0.0114$                 & $0.0768\pm 0.0012$ & $0.0781\pm 0.0043$ \\ 
                      & \ourmethod           & $0.0289\pm 0.005$                 & $0.0782\pm 0.004$ & $0.0791\pm 0.0071$ \\ 
\midrule
\electric             & \ourmethod+           & $0.0144\pm 0.0025$                 & $0.0971\pm 0.0039$ &  $0.103\pm 0.0055$ \\
                      & \ourmethod           & $0.0161\pm 0.0020$                 & $0.0861\pm 0.0013$ &  $0.0876\pm 0.0019$ \\
\bottomrule
\end{tabular}
}
\end{table}

\subsection{Results when using other base forecasters}

In addition to the seq2seq model, we also employ a Theta model and an Informer for base forecasting. We then apply the baselines and \ourmethod to calibrate the forecasting results. The results are shown in Table \ref{table: other base forecaster}. When using the Theta model, our method is not the best on \weather. When using the Informer, we are the second place on \smd. However, \ourmethod still outperforms all the baselines in calibration score except the two cases mentioned above and is close to the best in CRPS and WIS.

\begin{table*}[htbp]
\centering
\caption{Performance metrics across different datasets and methods using ThetaModel and Informer as the base forecaster. The shown results are for one step ahead predictions. Note that the baselines are deterministic, so only \ourmethod is averaged across four experiments. The standard deviations for the ThetaModel and Informer are shown in Table \ref{table: other base model std}.}
\label{table: other base forecaster}
\resizebox{0.99\linewidth}{!}{
\begin{tabular}{l|ccc|ccc|ccc|ccc|ccc}
\toprule
\textbf{ThetaModel} & \multicolumn{3}{c}{NCC} & \multicolumn{3}{c}{C-PID} & \multicolumn{3}{c}{ACI} & \multicolumn{3}{c}{NEXCP} & \multicolumn{3}{c}{CF-RNN} \\
\cmidrule(lr){1-1} \cmidrule(lr){2-4} \cmidrule(lr){5-7} \cmidrule(lr){8-10} \cmidrule(lr){11-13} \cmidrule(lr){14-16}
Datasets & CS & CRPS & WIS & CS & CRPS & WIS & CS & CRPS & WIS & CS & CRPS & WIS & CS & CRPS & WIS \\
\midrule
Covid & 0.0595 & 0.173 & 0.157 & 0.102 & 0.227 & 0.207      & 0.0841 & 0.176 & 0.163     & 0.143 & 0.174 & 0.160 & 0.109 & 0.175 & 0.163 \\ \hline
Flu   & 0.0678 & 0.162 & 0.160 & 0.0938 & 0.266 & 0.498     & 0.0813 & 0.154 & 0.147     & 0.0821 & 0.153 & 0.145   & 0.0912 & 0.154 & 0.146 \\ \hline
SMD   & 0.0191 & 0.0346 & 0.0361 & 0.0486 & 0.0252 & 0.0279   & 0.0191 & 0.0241 & 0.0231   & 0.0275 & 0.0240 & 0.0229   & 0.0310 & 0.0240 & 0.0230 \\  \hline
Elec  & 0.0173 & 0.0676 & 0.0698 & 0.0475 & 0.0584 & 0.0630   & 0.0347 & 0.0587 & 0.0554   & 0.0352 & 0.0584 & 0.0552 & 0.0402 & 0.0587 & 0.0552 \\ \hline
Weather & 0.0413 & 0.0466 & 0.0346 & 0.0382 & 0.0370 & 0.0368   & 0.0291 & 0.0345 & 0.0327   & 0.0520 & 0.0347 & 0.0330   & 0.0680 & 0.0347 & 0.0328 \\ 
\bottomrule
\end{tabular}}
\\
\vspace{0.3cm}
\resizebox{0.99\linewidth}{!}{
\begin{tabular}{l|ccc|ccc|ccc|ccc|ccc}
\toprule
\textbf{Informer} & \multicolumn{3}{c}{NCC} & \multicolumn{3}{c}{C-PID} & \multicolumn{3}{c}{ACI} & \multicolumn{3}{c}{NEXCP} & \multicolumn{3}{c}{CF-RNN} \\
\cmidrule(lr){1-1} \cmidrule(lr){2-4} \cmidrule(lr){5-7} \cmidrule(lr){8-10} \cmidrule(lr){11-13} \cmidrule(lr){14-16}
Datasets & CS & CRPS & WIS & CS & CRPS & WIS & CS & CRPS & WIS & CS & CRPS & WIS & CS & CRPS & WIS \\
\midrule
Covid & 0.0519 & 0.0711  & 0.0573  & 0.0997 & 0.0623 & 0.0787 & 0.0607 & 0.0554 & 0.0514 & 0.0615 & 0.0543 & 0.0505 & 0.0743 & 0.0551 & 0.0510 \\ \hline
Flu   & 0.0987  & 0.106    & 0.103 & 0.117 & 0.127 & 0.194    & 0.106 & 0.101 & 0.0971   & 0.109 & 0.101 & 0.0963   & 0.135 & 0.102 & 0.0969 \\ \hline
SMD   & 0.0551  & 0.121   & 0.120  & 0.0548 & 0.110 & 0.101   & 0.0453 & 0.115 & 0.104   & 0.0560 & 0.114 & 0.102   & 0.0775 & 0.116 & 0.105 \\  \hline
Elec  & 0.0222  & 0.0777  & 0.0808 & 0.0469 & 0.0712 & 0.0732 & 0.0319 & 0.0714 & 0.0677 & 0.0430 & 0.0710 & 0.0674 & 0.0375 & 0.0714 & 0.0675 \\ \hline
Weather & 0.0244  & 0.134  & 0.129 & 0.130 & 0.131 & 0.0433   & 0.0323 & 0.134 & 0.126   & 0.0585 & 0.132 & 0.125   & 0.0633 & 0.134 & 0.127 \\ 
\bottomrule
\end{tabular}}
\end{table*}

\begin{table}[htbp]
\centering
\caption{Standard deviation of \ourmethod on four experiments using the same set of hyperparameters with different seeds when using ThetaModel and Informer as the base forecaster.}
\label{table: other base model std}
\resizebox{0.99\linewidth}{!}{
\begin{tabular}{l|ccc|ccc}
\toprule
Base forecaster & \multicolumn{3}{c}{ThetaModel} & \multicolumn{3}{c}{Informer}\\
Datasets & CS & CRPS & WIS & CS & CRPS & WIS  \\
\cmidrule(lr){1-1} \cmidrule(lr){2-4} \cmidrule(lr){5-7}
Covid & 0.0595 & 0.173 & 0.157         & 0.0040 & 0.0030 & 0.0006            \\ \hline
Flu   & 0.0678 & 0.162 & 0.160         & 0.0024 & 0.001 & 0.001            \\ \hline
SMD   & 0.0191 & 0.0346 & 0.0361       & 0.0245 & 0.003 & 0.003            \\ \hline
Elec  & 0.0173 & 0.0676 & 0.0698       & 0.00061 & 0.0018 & 0.0030           \\ \hline
Weather & 0.0413 & 0.0466 & 0.0346     & 0.0020 & 0.0023 & 0.0006            \\ 
\bottomrule
\end{tabular}}
\end{table}

\subsection{More on the results}
The numerical results of Figure \ref{figure: wis_results} are provided in Table \ref{table: numerical results std}.

\begin{table*}[htbp]
\centering
\caption{Performance metrics across different datasets and methods using a Seq2seq model as the base forecaster. Note that the baselines are deterministic, so only \ourmethod is averaged across four experiments. The standard deviation is shown in Table \ref{table: numerical results std}.}
\label{table: numerical results}
\resizebox{0.99\linewidth}{!}{
\begin{tabular}{lc|ccc|ccc|ccc|ccc|ccc}
\toprule
Methods & & \multicolumn{3}{c}{NCC} & \multicolumn{3}{c}{C-PID} & \multicolumn{3}{c}{ACI} & \multicolumn{3}{c}{NEXCP} & \multicolumn{3}{c}{CF-RNN} \\
\cmidrule(lr){3-5} \cmidrule(lr){6-8} \cmidrule(lr){9-11} \cmidrule(lr){12-14} \cmidrule(lr){15-17}
Datasets & Step ahead & CS & CRPS & WIS & CS & CRPS & WIS & CS & CRPS & WIS & CS & CRPS & WIS & CS & CRPS & WIS \\
\midrule
Covid & 1 & 0.0551 & 0.116 & 0.113 & 0.1140 & 0.181 & 0.281 & 0.0755 & 0.123 & 0.118 & 0.0722 & 0.121 & 0.114 & 0.0798 & 0.122 & 0.117 \\
 & 2 & 0.0728 & 0.158 & 0.151 & 0.0970 & 0.169 & 0.190 & 0.0969 & 0.162 & 0.158 & 0.103 & 0.159 & 0.153 & 0.109 & 0.163 & 0.158 \\
 & 3 & 0.0892 & 0.196 & 0.188 & 0.0932 & 0.212 & 0.255 & 0.111 & 0.196 & 0.191 & 0.125 & 0.194 & 0.186 & 0.128 & 0.196 & 0.189 \\
 & 4 & 0.0971 & 0.223 & 0.213 & 0.103 & 0.801 & 1.98 & 0.113 & 0.227 & 0.221 & 0.117 & 0.226 & 0.217 & 0.130 & 0.227 & 0.217 \\ \hline
Flu & 1 & 0.104 & 0.142 & 0.134    & 0.171 & 0.123 & 0.120    & 0.109 & 0.120 & 0.116    & 0.106 & 0.118 & 0.113    & 0.125 & 0.120 & 0.115 \\
    & 2 & 0.0984 & 0.144 & 0.134    & 0.165 & 0.143 & 0.140    & 0.122 & 0.142 & 0.138    & 0.120 & 0.139 & 0.134    & 0.137 & 0.141 & 0.137 \\
    & 3 & 0.103 & 0.149 & 0.141    & 0.164 & 0.163 & 0.165    & 0.131 & 0.163 & 0.161    & 0.133 & 0.161 & 0.156    & 0.147 & 0.163 & 0.158 \\
    & 4 & 0.105 & 0.160 & 0.150    & 0.161 & 0.187 & 0/195    & 0.138 & 0.187 & 0.186    & 0.145 & 0.185 & 0.180    & 0.160 & 0.187 & 0.182 \\ \hline
SMD & 1 & 0.0289 & 0.0782 & 0.0791 & 0.0370 & 0.0680 & 0.0630 & 0.0346 & 0.0692 & 0.0656 & 0.0453 & 0.0677 & 0.0632 & 0.0632 & 0.0696 & 0.0665 \\ 
 & 2 & 0.0399 & 0.0946 & 0.0966 & 0.0677 & 0.0833 & 0.0958 & 0.0465 & 0.0863 & 0.0826 & 0.0644 & 0.0848 & 0.0799 & 0.0810 & 0.0865 & 0.0830 \\
 & 3 & 0.0304 & 0.0963 & 0.0984 & 0.0796 & 0.0907 & 0.106 & 0.0644 & 0.0930 & 0.0884 & 0.0982 & 0.0916 & 0.0868 & 0.100 & 0.0931 & 0.0892 \\
 & 4 & 0.0417 & 0.106 & 0.107 & 0.0236 & 0.0953 & 0.114 & 0.0655 & 0.0951 & 0.0909 & 0.0822 & 0.0938 & 0.0899 & 0.0775 & 0.0953 & 0.0920 \\ \hline
Elec & 1 & 0.0161 & 0.0861 & 0.0876 & 0.0469 & 0.0783 & 0.0876 & 0.0275 & 0.0810 & 0.0766 & 0.0452 & 0.0800 & 0.0757 & 0.0530 & 0.0869 & 0.0769 \\ 
 & 3 & 0.0237 & 0.102 & 0.103 & 0.0769 & 0.124 & 0.205 & 0.0369 & 0.0968 & 0.0938 & 0.0758 & 0.0957 & 0.0928 & 0.0875 & 0.0964 & 0.0930 \\
 & 6 & 0.0297 & 0.0975 & 0.0937 & 0.125 & 0.176 & 0.331 & 0.0424 & 0.0992 & 0.0967 & 0.0780 & 0.0979 & 0.0956 & 0.0875 & 0.0986 & 0.0955 \\
 & 12 & 0.0347 & 0.114 & 0.111 & 0.157 & 0.249 & 0.549 & 0.0513 & 0.119 & 0.121 & 0.0852 & 0.117 & 0.116 & 0.0941 & 0.119 & 0.117 \\ \hline
Weather & 1 & 0.0221 & 0.115 & 0.108 & 0.0433 & 0.122 & 0.125 & 0.0647 & 0.126 & 0.120 & 0.0693 & 0.124 & 0.116 & 0.0636 & 0.127 & 0.120 \\ 
 & 3 & 0.0560 & 0.187 & 0.184 & 0.0787 & 0.192 & 0.240 & 0.0693 & 0.185 & 0.180 & 0.0682 & 0.183 & 0.178 & 0.0624 & 0.185 & 0.179 \\
 & 5 & 0.0527 & 0.197 & 0.195 & 0.0984 & 0.234 & 0.393 & 0.0751 & 0.197 & 0.193 & 0.0990 & 0.196 & 0.193 & 0.0758 & 0.195 & 0.191 \\
 & 10 & 0.0297 & 0.388 & 0.362 & 0.144 & 0.947 & 1.97 & 0.142 & 0.442 & 0.441 & 0.186 & 0.439 & 0.451 & 0.174 & 0.439 & 0.442 \\
\bottomrule
\end{tabular}}
\end{table*}

\begin{table}[htbp]
\centering
\caption{Standard deviation of \ourmethod on four experiments using the same set of hyperparameters with different seeds when using a seq2seq model as the base forecaster.}
\label{table: numerical results std}
\resizebox{0.66\linewidth}{!}{
\begin{tabular}{lc|ccc}
\toprule
Methods & & \multicolumn{3}{c}{NCC} \\
\cmidrule(lr){3-5}
Datasets & Step ahead & CS & CRPS & WIS \\
\midrule
Covid & 1 & 0.0028 & 0.001 & 0.001 \\
 & 2 & 0.0083 & 0.001 & 0.002 \\
 & 3 & 0.0100 & 0.007 & 0.008 \\
 & 4 & 0.0020 & 0.009 & 0.002 \\ \hline
Flu & 1 & 0.004 & 0.010 & 0.010     \\
    & 2 & 0.0018 & 0.002 & 0.001    \\
    & 3 & 0.005 & 0.001 & 0.002     \\
    & 4 & 0.007 & 0.002 & 0.002     \\ \hline
SMD & 1 & 0.0039 & 0.0045 & 0.0070 \\ 
    & 2 & 0.0111 & 0.0025 & 0.0038 \\
    & 3 & 0.0062 & 0.0044 & 0.0052 \\
    & 4 & 0.0070 & 0.009 & 0.013   \\ \hline
Elec & 1 & 0.0016 & 0.0010 & 0.0013  \\ 
 & 3 & 0.0054 & 0.001 & 0.002        \\
 & 6 & 0.0082 & 0.0035 & 0.0030      \\
 & 12 & 0.0056 & 0.002 & 0.003       \\ \hline
Weather & 1 & 0.0051 & 0.115 & 0.001 \\ 
 & 3 & 0.0111 & 0.009 & 0.015        \\
 & 5 & 0.0051 & 0.006 & 0.009        \\
 & 10 & 0.0014 & 0.001 & 0.001       \\
\bottomrule
\end{tabular}}
\end{table}

\subsection{Effect of sorting on C-PID}
\begin{figure}[!htb]
\minipage{0.49\linewidth}
  \centering
  \includegraphics[width=\linewidth]{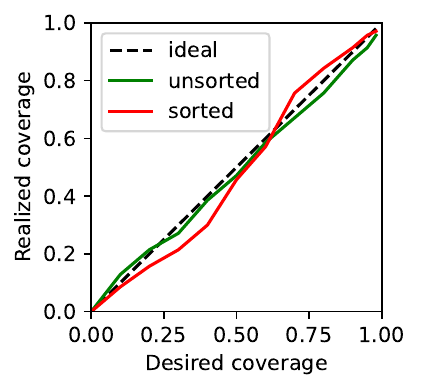}
  \text{(a)}
\endminipage\hfill
\minipage{0.49\linewidth}
  \centering
  \includegraphics[width=\linewidth]{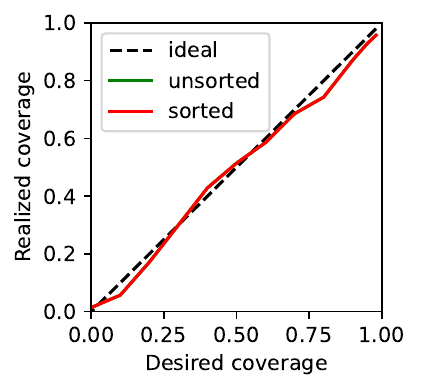}
  \text{(b)}
\endminipage\hfill
\caption{Calibration curve of unsorted and sorted results on \smd. (a) shows results from C-PID. (b) shows results from \ourmethod.}
\label{figure: pid cs}
\end{figure}

From the main results, we can see that C-PID often yields poor results. However, in Figure \ref{figure: combined}(a), C-PID achieves the best calibration scores on most datasets before sorting. This raises the question: how much influence does sorting have on C-PID's calibration performance? In Figure \ref{figure: pid cs}, we plot the calibration curve of both the sorted and unsorted results on the \smd dataset. For unsorted results of C-PID shown in Figure \ref{figure: pid cs}.a, the empirical coverage curve closely follows the ideal one. However, after sorting, the curve below 0.5 shifts downward, and the curve above 0.5 shifts upward, indicating a significant decline in calibration performance. We also show the calibration curve for \ourmethod in Figure \ref{figure: pid cs}.b, from which we can see that the empirical coverage remains close to the ideal curve for both the sorted and unsorted version.

\subsection{Qualitative Comparison}
To further compare the prediction intervals generated by \ourmethod with the baselines, we plot the 90\% prediction interval along with the ground truth and base model predictions on \weather dataset. 

\begin{figure}
    \centering
    \includegraphics[width=0.99\linewidth]{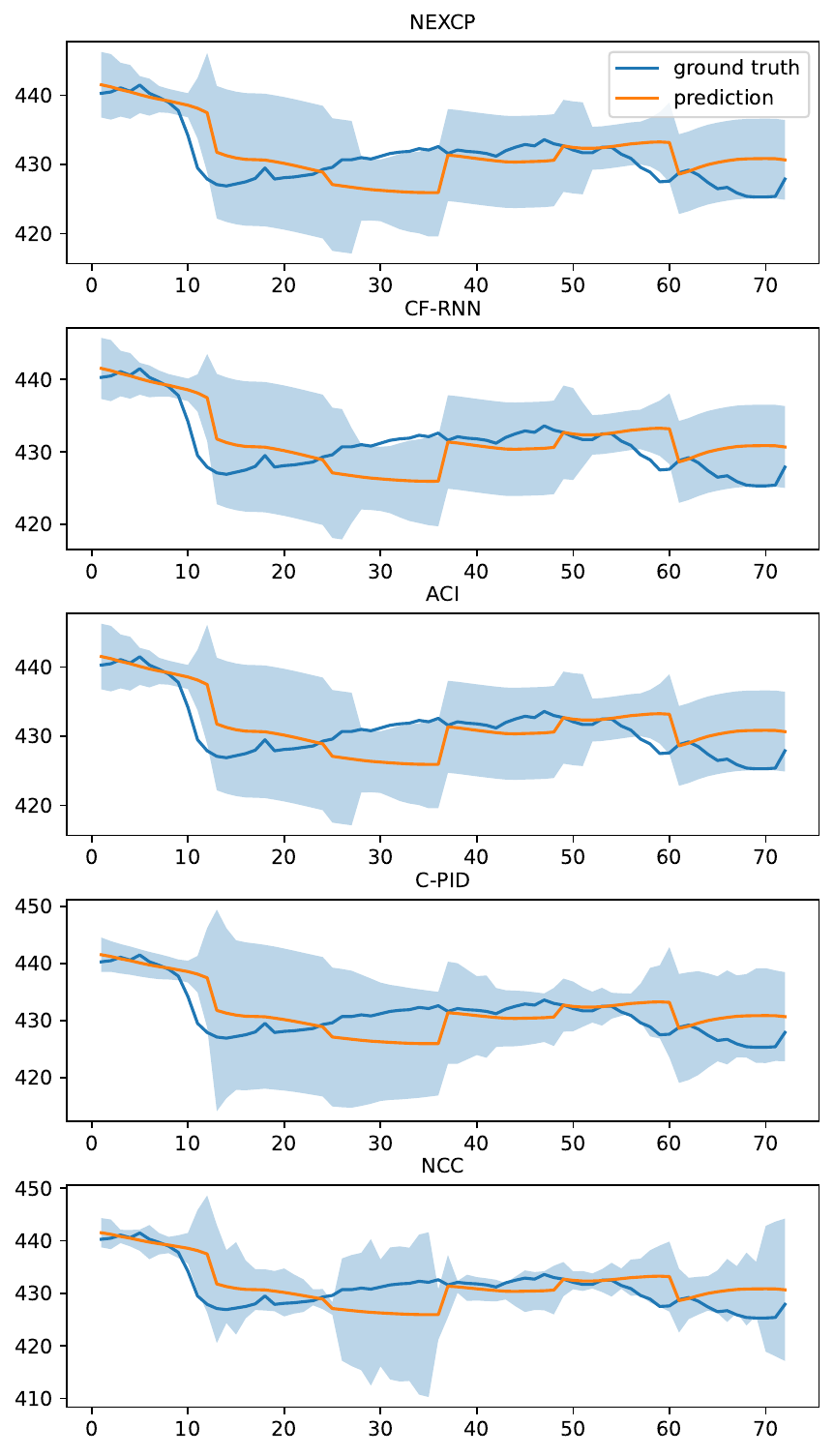}
    \caption{90\% prediction intervals on \weather dataset. The orange line shows the predictions. The blue line shows the ground truth data. The blue area shows the 90\% prediction intervals.}
    \label{figure: 90pil}
\end{figure}

\newpage

\section{Reproducibility Checklist}
Unless specified otherwise, please answer “yes” to each question if the relevant information is described either in the paper itself or in a technical appendix with an explicit reference from the main paper. If you wish to explain an answer further, please do so in a section titled “Reproducibility Checklist” at the end of the technical appendix. \\

\par\noindent This paper:
\begin{itemize}
    \item Includes a conceptual outline and/or pseudocode description of AI methods introduced (\textbf{yes})
    \item Clearly delineates statements that are opinions, hypothesis, and speculation from objective facts and results (\textbf{yes})
    \item Provides well marked pedagogical references for less-familiare readers to gain background necessary to replicate the paper (\textbf{yes}) \\
\end{itemize}

\par\noindent Does this paper make theoretical contributions? (\textbf{yes}) \\

\par\noindent If yes, please complete the list below.
\begin{itemize}
    \item All assumptions and restrictions are stated clearly and formally. (\textbf{yes})
    \item All novel claims are stated formally (e.g., in theorem statements). (\textbf{yes})
    \item Proofs of all novel claims are included. (\textbf{yes})
    \item Proof sketches or intuitions are given for complex and/or novel results. (\textbf{yes})
    \item Appropriate citations to theoretical tools used are given. (\textbf{yes})
    \item All theoretical claims are demonstrated empirically to hold. (\textbf{yes})
    \item All experimental code used to eliminate or disprove claims is included. (\textbf{yes}) \\
\end{itemize} 

\par\noindent Does this paper rely on one or more datasets? (\textbf{yes}) \\

\par\noindent If yes, please complete the list below.
\begin{itemize}
    \item A motivation is given for why the experiments are conducted on the selected datasets (\textbf{yes})
    \item All novel datasets introduced in this paper are included in a data appendix. (\textbf{yes})
    \item All novel datasets introduced in this paper will be made publicly available upon publication of the paper with a license that allows free usage for research purposes. (\textbf{yes})
    \item All datasets drawn from the existing literature (potentially including authors’ own previously published work) are accompanied by appropriate citations. (\textbf{yes})
    \item All datasets drawn from the existing literature (potentially including authors’ own previously published work) are publicly available. (\textbf{yes})
    \item All datasets that are not publicly available are described in detail, with explanation why publicly available alternatives are not scientifically satisficing. (\textbf{yes}) \\
\end{itemize}

\par\noindent Does this paper include computational experiments? (\textbf{yes}) \\

\par\noindent If yes, please complete the list below.
\begin{itemize}
    \item Any code required for pre-processing data is included in the appendix. (\textbf{yes}).
    \item All source code required for conducting and analyzing the experiments is included in a code appendix. (\textbf{yes})
    \item All source code required for conducting and analyzing the experiments will be made publicly available upon publication of the paper with a license that allows free usage for research purposes. (\textbf{yes})
    \item All source code implementing new methods have comments detailing the implementation, with references to the paper where each step comes from (\textbf{yes})
    \item If an algorithm depends on randomness, then the method used for setting seeds is described in a way sufficient to allow replication of results. (\textbf{yes})
    \item This paper specifies the computing infrastructure used for running experiments (hardware and software), including GPU/CPU models; amount of memory; operating system; names and versions of relevant software libraries and frameworks. (\textbf{yes})
    \item This paper formally describes evaluation metrics used and explains the motivation for choosing these metrics. (\textbf{yes})
    \item This paper states the number of algorithm runs used to compute each reported result. (\textbf{yes})
    \item Analysis of experiments goes beyond single-dimensional summaries of performance (e.g., average; median) to include measures of variation, confidence, or other distributional information. (\textbf{yes})
    \item The significance of any improvement or decrease in performance is judged using appropriate statistical tests (e.g., Wilcoxon signed-rank). (\textbf{yes})
    \item This paper lists all final (hyper-)parameters used for each model/algorithm in the paper’s experiments. (\textbf{Yes})
    \item This paper states the number and range of values tried per (hyper-) parameter during development of the paper, along with the criterion used for selecting the final parameter setting. (\textbf{Yes})
\end{itemize}

\end{document}